\documentclass[twoside,11pt]{article}

\usepackage[preprint]{jmlr2e}
\usepackage{comment}
\usepackage{wrapfig}
\usepackage{amsmath}
\usepackage{enumitem}
\usepackage{booktabs}
\usepackage{algorithm}
\usepackage{algorithmic}

\newcommand{\archmath}{a}

\newcommand{\searchspace}{$\mathcal{A}$ }
\newcommand{\searchspacemath}{\mathcal{A}}

\newcommand{\lval}{\mathcal{L}_{\text{val}}}
\newcommand{\ltrain}{\mathcal{L}_{\text{train}}}

\newcommand{\nasobjective}{$f(a)$ }
\newcommand{\nasobjectivemath}{f}

\DeclareMathOperator*{\argmin}{arg\,min}

\newcommand{\archss}{\mathcal{A}}

\newcommand{\ens}{\small{\texttt{Ensemble}}}

\usepackage{pifont}
\newcommand{\cmark}{\ding{51}}%
\usepackage{lastpage}
\jmlrheading{23}{2022}{1-\pageref{LastPage}}{1/21; Revised 5/22}{9/22}{21-0000}{Colin White, Mahmoud Safari, Rhea Sukthanker, Binxin Ru, Thomas Elsken, Arber Zela, Debadeepta Dey and Frank Hutter}

\ShortHeadings{Neural Architecture Search: Insights from 1000 Papers}{White, Safari, Sukthanker, Ru, Elsken, Zela, Dey and Hutter}
\firstpageno{1}

\begin{document}

\title{Neural Architecture Search: Insights from 1000 Papers}

\author{\name Colin White \email colin@abacus.ai \\
       \addr Abacus.AI\\
       San Francisco, CA 94105, USA
       \AND
       \name Mahmoud Safari \email safarim@cs.uni-freiburg.de \\
       \addr University of Freiburg\\
       Freiburg im Breisgau, 79110, Germany
       \AND
       \name Rhea Sukthanker \email sukthank@cs.uni-freiburg.de \\
       \addr University of Freiburg\\
       Freiburg im Breisgau, 79110, Germany
       \AND
       \name Binxin Ru \email robinru@sailyond.com \\
       \addr Sailyond Technology \& Research Institute of Tsinghua University \\
        Shenzhen, 518071, China
       \AND
       \name Thomas Elsken \email thomas.elsken@de.bosch.com \\
       \addr Bosch Center for Artificial Intelligence\\
       Renningen, 71272, Germany
       \AND
       \name Arber Zela \email zelaa@cs.uni-freiburg.de \\
       \addr University of Freiburg\\
       Freiburg im Breisgau, 79110, Germany
       \AND
       \name Debadeepta Dey \email dedey@microsoft.com \\
       \addr Microsoft Research\\
       Redmond, WA 98052, USA
       \AND
       \name Frank Hutter \email fh@cs.uni-freiburg.de \\
       \addr University of Freiburg \& Bosch Center for Artificial Intelligence\\
       Freiburg im Breisgau, 79110, Germany}

\editor{My editor}

\maketitle

\begin{abstract}
In the past decade, advances in deep learning have resulted in breakthroughs in a variety of areas, including computer vision, natural language understanding, speech recognition, and reinforcement learning. Specialized, high-performing neural architectures are crucial to the success of deep learning in these areas. Neural architecture search (NAS), the process of automating the design of neural architectures for a given task, is an inevitable next step in automating machine learning and has already outpaced the best human-designed architectures on many tasks. In the past few years, research in NAS has been progressing rapidly, with over 1000 papers released since 2020 \citep{deng-21}.
In this survey, we provide an organized and comprehensive guide to neural architecture search. We give a taxonomy of search spaces, algorithms, and speedup techniques, and we discuss resources such as benchmarks, best practices, other surveys, and open-source libraries.
\end{abstract}


\section{Introduction} \label{sec:intro}
In the past decade, deep learning has become the dominant paradigm in machine learning for a variety of applications and has been used in a number of breakthroughs across computer vision 
\citep{alexnet, Huang_2017_CVPR, szegedy2017inception,he2016deep}, 
natural language understanding \citep{vaswani2017attention, hochreiter1997long, bahdanau2014neural}, 
speech recognition \citep{hannun2014deep, chorowski2015attention, chan2016listen}, and
reinforcement learning \citep{Mnih2015, silver2016mastering}; it is also becoming a very powerful approach for the analysis of
tabular data \citep{kadra2021regularization, saint,tabpfn}.
While many factors played into the rise of deep learning approaches, including deep learning's ability to automate feature extraction, as well as an increase in data and the larger availability of computational resources,
the design of high-performing neural architectures has been crucial to the success of deep learning.
Recently, just as manual feature engineering was replaced by automated feature learning via deep learning,
it is getting more and more common to automate the time-consuming architecture design step via 
\emph{neural architecture search}.
Neural architecture search (NAS), the process of automating the design of neural architectures for a given task, has already outpaced the best human-designed architectures on many tasks \citep{zoph2018learning,dpc, Ghiasi_2019_CVPR, Du_2020_CVPR, so2019evolved},
notably ImageNet \citep{hu2019efficient,real2019regularized,zoph2018learning,pnas}, as well as diverse and less-studied datasets \citep{shen2022efficient}, and in memory- or latency-constrained settings \citep{benmeziane2021comprehensive}.
Indeed, in the past few years, research in NAS has been progressing rapidly.
Although several surveys have been written for NAS and related areas in the past 
\citep[also see Section \ref{subsec:surveys}]{nas-survey, wistuba2019survey},
\textbf{over 1000 new NAS papers have been released in the last two years} \citep{deng-21}, 
warranting the need for a new survey on over-arching advances, which we aim to provide with this work.

\subsection{A Brief History of NAS and Relation to Other Fields}

NAS emerged as a subfield of \emph{automated machine learning (AutoML)} \citep{automl}, the process of automating all steps in the machine learning pipeline, from data cleaning, to feature engineering and selection, to hyperparameter and architecture search. 
NAS has a large overlap with \emph{hyperparameter optimization (HPO)} \citep{feurer_hyperparameter_2019}, which refers to the automated optimization of hyperparameters of the machine learning model.
NAS is sometimes referred to as a subset of HPO \citep{randomnas}, since NAS can be expressed as optimizing only the hyperparameters that correspond to the architecture, a subset of the entire set of model hyperparameters. However, the techniques for HPO vs.\ NAS are often substantially different.

A typical HPO problem optimizes a mix of continuous and categorical hyperparameters, such as learning rate, dropout rate, batch size, momentum, activation function, normalization strategy, and so on.
Typically, the domains of most hyperparameters are independent (that is, the set of possible values for each hyperparameter is not affected by the possible values of other hyperparameters).
Therefore, the typical \emph{search space} of an HPO problem is the product space of a mix of continuous and categorical dimensions.
By contrast, NAS is specifically focused on optimizing the \emph{topology of the architecture}, which can be much more complex. 
The topology is typically represented by a directed acyclic graph (DAG), in which the nodes or edges are labeled by neural network operations.
Therefore, the \emph{search space} of a NAS problem is typically discrete%
\footnote{
Notably, some NAS techniques such as DARTS \citep{darts} relax the domain to be continuous during the search, but then the hyperparameters are discretized in order to return the final architecture.}
and can be represented directly as a graph, or as a hierarchical structure of conditional hyperparameters.

Although standard HPO algorithms can sometimes be adapted for NAS \citep{mendoza-automl16a, zela2018towards,zimmer-tpami21a,izquierdo2021bag,abohb,li2018system}, 
it is often much more efficient and effective to use NAS techniques which are tailored to optimize the intricate space of neural architectures.
Furthermore, most modern NAS techniques go beyond black-box optimization algorithms by exploiting details specific to NAS, such as sharing weights among similar neural architectures to avoid training each of them from scratch.

\begin{wrapfigure}{t}{3.0in} 
\vspace{-6mm}
\begin{minipage}{3.0in}
\centering
\resizebox{3.0in}{!}{\includegraphics{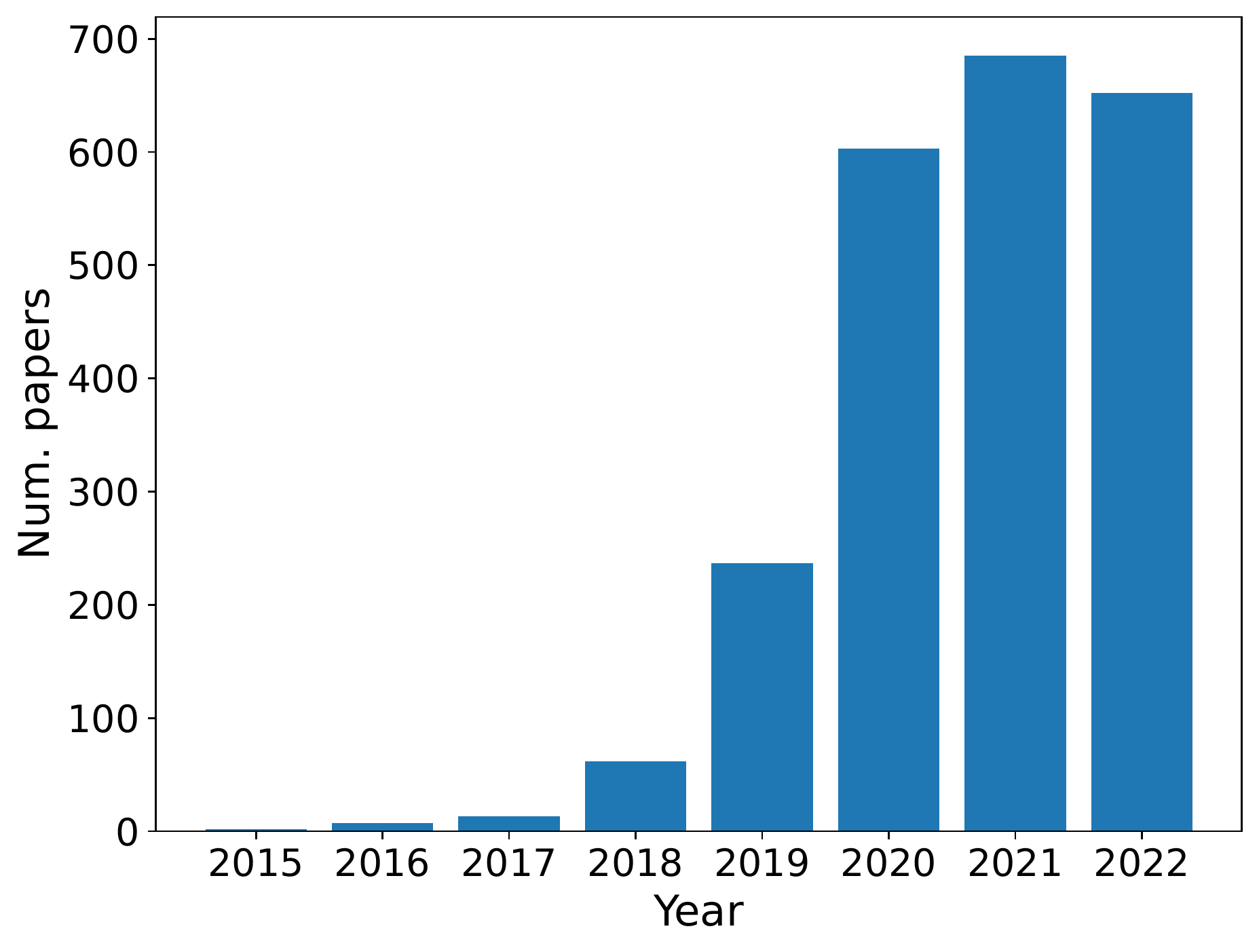}}
\vspace{-7mm}
\caption{Number of NAS papers by year \citep{deng-21}.
}
\vspace{-6mm}
\label{fig:nas-paper-graph}
\end{minipage}
\end{wrapfigure}

%

Historically, NAS has been around since at least the late 1980s \citep{tenorio1988self, miller1989designing, kitano1990designing, angeline1994evolutionary} but it did not gain widespread attention until the popular paper, \emph{NAS with Reinforcement Learning}, by \citet{zoph2017neural}. 
There has since been a huge interest in NAS, with over 1000 papers released in the last two years (see Figure \ref{fig:nas-paper-graph}).

By now, many different approaches, such as reinforcement learning, evolutionary algorithms, Bayesian optimization, and NAS-specific techniques based on weight sharing have been explored. 
Perhaps the most popular recent approaches are one-shot techniques \citep{bender2018understanding, darts}, which often substantially speed up the search process compared to black-box optimization techniques.
In recent years, a large body of follow-up work has focused on making one-shot methods more robust and reliable \citep{ zela2020understanding, wang2021rethinking}.
In parallel, there has been a large push to make NAS research more reproducible and scientific,
starting with the release of NAS-Bench-101 \citep{nasbench}, the first tabular benchmark for NAS. 
Furthermore, while the early days of NAS has mostly focused on image classification problems such as CIFAR-10 and ImageNet, the field has now expanded to many other domains, such as 
object detection \citep{Ghiasi_2019_CVPR,Xu_2019_ICCV}, 
semantic segmentation \citep{dpc,Liu_2019_CVPR}, 
speech recognition \citep{nasbenchasr}, 
partial differential equation solving \citep{nasbench360,roberts2021rethinking,shen2022efficient}, 
protein folding \citep{roberts2021rethinking,shen2022efficient}, 
and weather prediction \citep{tu2022automl}, 
and the field has seen a renewed interest in natural language processing \citep{chitty2022neural,javaheripi2022litetransformersearch}.

\subsection{Background and Definitions}

\begin{figure}
    \centering
    \includegraphics[width=0.9\linewidth]{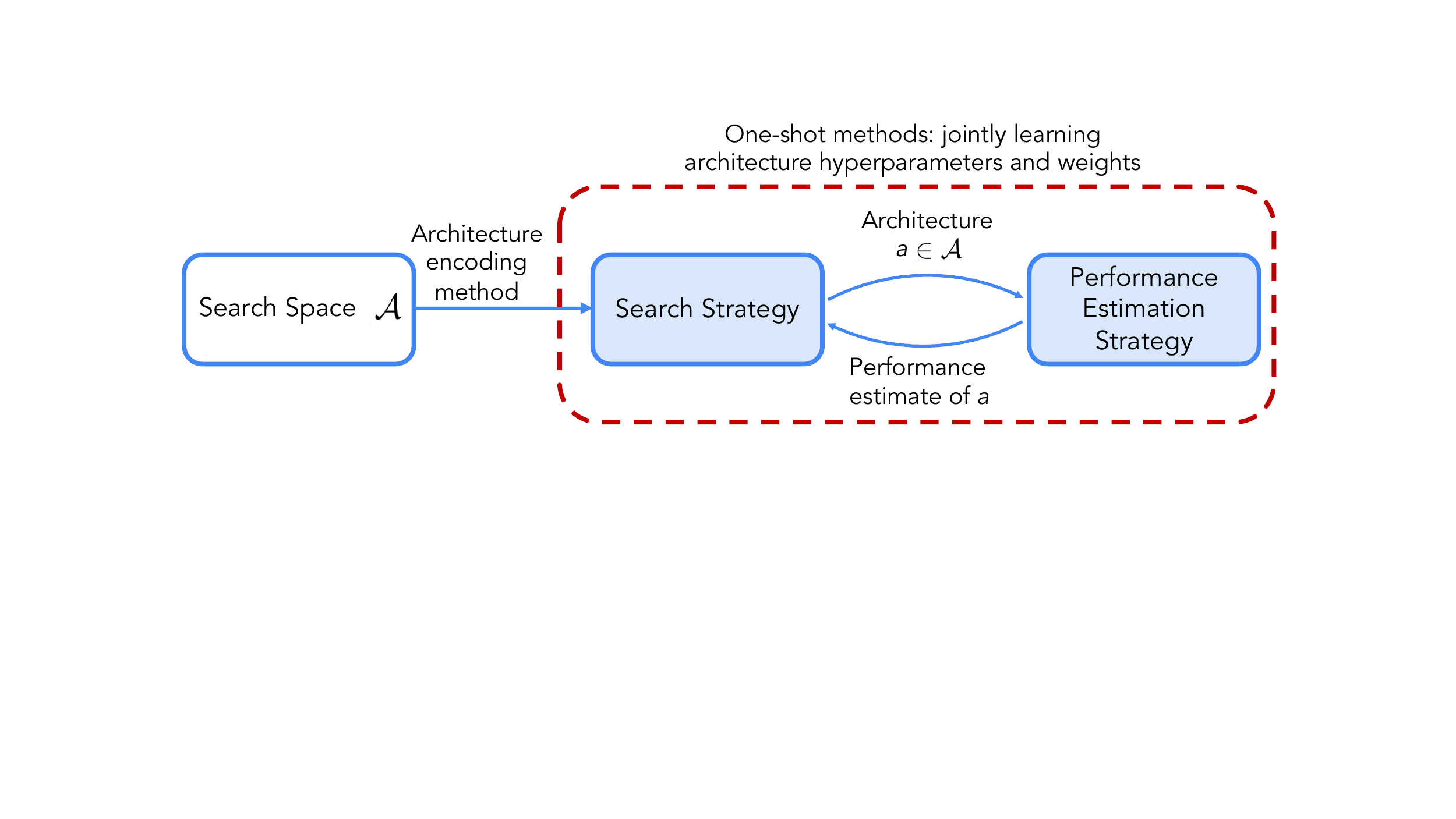}
    \vspace{-4mm}
    \caption{Overview of neural architecture search \citep{nas-survey,weng2020neural}.
    A search strategy iteratively selects architectures (typically by using an architecture encoding method) from a predefined search space $\archss{}$.
    The architectures are passed to a performance estimation strategy, which returns the performance estimate to the search strategy.
    For \emph{one-shot methods}, the search strategy and performance estimation strategy are inherently coupled.
    }
    \vspace{-3mm}
    \label{fig:overview}
\end{figure}

Prior NAS surveys \citep[e.g.][]{nas-survey, wistuba2019survey} have referred to three dimensions of NAS: 
\emph{search space, search strategy}, and \emph{performance evaluation strategy} (see Figure \ref{fig:overview}).
We define each term below, as this is a useful disambiguation for understanding many NAS methods. However, it is worth noting that the trichotomy cannot be applied to the large sub-area of one-shot methods, because 
for these methods, the search strategy is coupled with the performance evaluation strategy \citep{xie2021weight}.

A \emph{search space} is the set of all architectures that the NAS algorithm is allowed to select. Common NAS search spaces range in size from a few thousand to over $10^{20}$.
While the search space in principle can be extremely general,
incorporating domain knowledge when designing the search space can simplify the search. 
However, adding too much domain knowledge introduces human bias, which reduces the chances of a NAS method finding truly novel architectures.
Search spaces are discussed in more detail in Section \ref{sec:search_spaces}.

A \emph{search strategy} is an optimization technique used to find a high-performing architecture in the search space.
There are generally two main categories of search strategies: black-box optimization based techniques (including multi-fidelity techniques) and one-shot techniques. 
However, there are some NAS methods for which both or neither category applies.
Black-box optimization based techniques, such as reinforcement learning, Bayesian optimization, and evolutionary search, are surveyed in Section \ref{sec:discrete}.
One-shot methods, including supernet- and hypernet-based methods, are surveyed in Section \ref{sec:oneshot}.

A \emph{performance estimation strategy} is any method used to quickly predict the performance of neural architectures in order to avoid fully training the architecture.
For example, while we can run a discrete search strategy by fully training and evaluating architectures chosen throughout the search, using a performance estimation strategy such as learning curve extrapolation can greatly increase the speed of the search.
Performance estimation strategies, and more generally speedup techniques, are surveyed in Section \ref{sec:speedup}.

The most basic definition of NAS is as follows.
Given a search space \searchspace, a dataset $\mathcal{D}$, a training pipeline $\mathcal{P}$, 
and a time or computation budget $t$,
the goal is to find an architecture $a\in \mathcal{A}$ within budget $t$ 
which has the highest possible validation accuracy when trained using dataset $\mathcal{D}$ and training pipeline $\mathcal{P}$.
A common method of approaching NAS is to approximately solve the following expression within time $t$:
\begin{equation*} \label{eq:bilevel}
    \min_{a\in\searchspacemath}\quad \lval\left(w^*(a),a\right)\quad
    \text{s.t.}\quad w^*(a) = \text{argmin}_w~ \ltrain\left(w,a\right).
\end{equation*}
Here, $\lval$ and $\ltrain$ denote the validation loss and training loss, respectively.
While this is the core definition of NAS, other variants will be discussed throughout this survey. 
For example, we may want to return an architecture with constraints on the number of parameters (Section \ref{subsec:hardware}), or we may use meta-learning (Section \ref{subsec:meta-learning}) to improve performance.

Throughout the rest of this article, we provide a comprehensive guide to the latest NAS techniques and resources.
Sections \ref{sec:search_spaces} to \ref{sec:speedup} are devoted to NAS techniques, surveying search spaces, black-box optimization techniques, one-shot techniques, and speedup techniques, respectively.
Sections \ref{sec:extensions} to \ref{sec:resources} cover extensions, applications, and resources, and Section \ref{sec:future_directions} concludes by discussing promising future directions.

\section{Search Spaces} \label{sec:search_spaces}

The search space is perhaps the most essential ingredient of NAS.
While other areas of AutoML overlap with NAS in terms of the optimization methods used, the architectural search space is unique to NAS.
Furthermore, the search space is often the first step when setting up NAS. 
The majority of popular search spaces are task-specific and were heavily inspired by the state-of-the-art manual architectures in their respective application domains. 
For example, NAS-Bench-101, a popular image classification search space 
\citep{nasbench} was inspired by ResNet \citep{he2016deep} and Inception \citep{szegedy2017inception}.

In fact, the design of the search space represents an important trade-off between human bias and efficiency of search: if the size of the search space is small and includes many hand-picked decisions, then NAS algorithms will have an easier time finding a high-performing architecture.
On the other hand, if the search space is large with more primitive building blocks, a NAS algorithm will need to run longer, but there is the possibility of discovering truly novel architectures \citep{real2020automl}.

In this section, we survey the main categories of search spaces for NAS as summarized in Table \ref{tab:search_space_summary}.
We start in Section \ref{subsec:ss_notation} by defining general terminology. In Sections \ref{subsec:macro} and \ref{subsec:chain}, we discuss the relatively simple macro and chain-structured search spaces, respectively.
In Section \ref{subsec:cell}, we describe the most popular type of search space: the cell-based search space. 
In Section \ref{subsubsec:hierarchical}, we describe hierarchical search spaces.
Finally, in Section \ref{subsec:encodings}, we discuss architecture encodings, an important design decision for NAS algorithms that is inherently tied to the choice of search space.

\subsection{Terminology} \label{subsec:ss_notation}
The search space terminologies differ across the literature, depending on the type of search space.
For clarity, we define the main terms here
and in Appendix Figure \ref{fig:search_space_terms}.

\begin{itemize}[topsep=0pt, itemsep=2pt, parsep=0pt, leftmargin=5mm]
    \item \emph{Operation/primitive} denotes the atomic unit of the search space.
    For nearly all popular search spaces, this is a triplet of a fixed activation, operation, and fixed normalization, such as \texttt{ReLU-conv\_1x1-batchnorm},
    where the \texttt{ReLU} and BatchNorm are fixed, and the middle operation is a choice among several different operations. 
    \item \emph{Layer} is often used in chain-structured or macro search spaces to denote the same thing as an operation or primitive. However, it sometimes refers to well-known combinations of operations, such as the \texttt{inverted bottleneck residual} \citep{sandler2018mobilenetv2,mnasnet, proxylessnas, tan2019efficientnet}.
    \item \emph{Block/Module} is sometimes used to denote a sequential stack of layers following the notation used in most chain-structured and macro search spaces \citep{ofa, mnasnet, tan2019efficientnet}. 
    \item \emph{Cell} is used to denote a directed acyclic graph of operations in cell-based search spaces. The maximum number of operations in a cell is often fixed.
    \item \emph{Motif} is used to denote a sub-pattern formed from multiple operations in an architecture. 
    Some literature refers to a cell as a higher-level motif and a smaller set of operations as a base-level motif.
\end{itemize}

\begin{table}[t] 
\resizebox{1.0\linewidth}{!}{
\begin{tabular}{@{}lllc@{}}
\toprule
\multicolumn{1}{c}{\textbf{Search Spaces}}  & \multicolumn{1}{c}{\begin{tabular}[c]{@{}c@{}} \textbf{Structure} \end{tabular}} & \multicolumn{1}{c}{\begin{tabular}[c]{@{}c@{}} \textbf{Searchable hyperparameters} \end{tabular}}                                                                 & \begin{tabular}[c]{@{}c@{}} \textbf{Levels of}  \\ \textbf{Topology}\end{tabular}  \\\midrule 
\begin{tabular}[c]{@{}l@{}} \textbf{Macro search space} \\ e.g. NASBOT \citep{nasbot},\\ EfficientNet  \citep{tan2019efficientnet} \end{tabular}  & \begin{tabular}[c]{@{}l@{}} DAG \end{tabular}                                                                                                & \begin{tabular}[c]{@{}l@{}} Operation types, DAG topology, \\ macro hyperparameters \end{tabular} & 1                          \\ \midrule
\begin{tabular}[c]{@{}l@{}} \textbf{Chain-structured search space} \\
e.g. MobileNetV2 \citep{sandler2018mobilenetv2} \end{tabular} & \begin{tabular}[c]{@{}l@{}} Chain \end{tabular}                              & Operation types, macro hyperparameters                                                                                                                                                                & 1                          \\\midrule
\begin{tabular}[c]{@{}l@{}} \textbf{Cell-based search space} \\ e.g. DARTS \citep{darts} \end{tabular}    & Duplicated cells                                                                                           & Operation type, cell topology                                                                                                                                                                   & 1                          \\\midrule
\begin{tabular}[c]{@{}l@{}} \textbf{Hierarchical search space} \\ e.g. Hier. Repr. \citep{liu2018hierarchical}, \\ Auto-DeepLab \citep{liu2019auto}  \end{tabular} & Varied                                                                                           & \begin{tabular}[c]{@{}l@{}}Operation type, cell/DAG topology, \\ macro hyperparameters \end{tabular} & $>1$ \\ \bottomrule
\end{tabular}}
\caption{Summary of the types of NAS search spaces.
\label{tab:search_space_summary}
}
\end{table}

\subsection{Macro Search Spaces} \label{subsec:macro}

In the NAS literature, macro search spaces may refer to one of two types. 
First, they may refer to search spaces which encode the entire architecture in one level (as opposed to cell-based or hierarchical search spaces), which were popular in 2017 and 2018.
Second, they may refer to search spaces which focus only on macro-level hyperparameters.

For the former, an entire architecture is represented as a single directed acyclic graph \citep{zoph2017neural, baker2016designing, Real17, nasbot}. 
These search spaces typically have a choice of operation at each node in the graph, as well as the choice of DAG topology.
For example, the NASBOT CNN search space \citep{nasbot} consists of choices of different convolution, pooling, and fully connected layers, with any DAG topology, with depth of at most 25.

The second type of macro search spaces \citep{tan2019efficientnet,natsbench,transnasbench}, focus on the variation of macro-level hyperparameters, such as where and how much to downsample the spatial resolution throughout the architecture, while keeping the architecture topology and operations fixed.%
\footnote{Strictly speaking, since these search spaces have a fixed architecture topology, they may also be called hyperparameter tuning search spaces instead of
NAS search spaces.}
For example, \citet{tan2019efficientnet} propose a CNN search space by varying the network depth, width, and input feature resolution.

Compared to other search spaces, macro search spaces have high representation power: their flexible structure allows the possibility of discovering novel architectures. 
However, their main downside is that they are very slow to search. 
In the next two sections, we discuss types of search spaces which have more rigidity, making them faster to search.

\subsection{Chain-Structured Search Spaces} \label{subsec:chain}

Chain-structured search spaces, as the name suggests, have a simple architecture topology: a sequential chain of operation layers.
They often take state-of-the-art manual designs, such as ResNet \citep{resnet} or MobileNets \citep{howard2017mobilenets}, as the backbone.

There are several chain-structured search spaces based on convolutional networks.
ProxylessNAS \citep{proxylessnas} starts with the MobileNetV2 \citep{sandler2018mobilenetv2} architecture and searches over the kernel sizes and expansion ratios in the inverted bottleneck residual layers. 
XD \citep{roberts2021rethinking} and DASH \citep{shen2022efficient} start with a LeNet \citep{lecun1999object}, ResNet \citep{he2016deep}, or WideResNet \citep{zagoruyko2016wide}, and search over an expressive generalization of convolutions based on Kaleidoscope matrices \citep{dao2019kaleidoscope}, or kernel sizes and dilations, respectively.

Chain-structured search spaces are also popular in transformer-based search spaces.
For example, the search space from Lightweight Transformer Search (LTS) \citep{javaheripi2022litetransformersearch} consists of a chain-structured configuration of the popular GPT family of architectures \citep{gpt2,gpt3} for autoregressive language modeling, with searchable choices for the number of layers, model dimension, adaptive embedding dimension, dimension of the feedforward neural network in a transformer layer, and number of heads in each transformer layer. 
The search spaces from NAS-BERT \citep{nasbert} and MAGIC \citep{xu2022analyzing} both consist of a chain-structured search space over the BERT architecture \citep{bert} with up to 26 operation choices consisting of variants of multi-head attention, feedforward layers, and convolutions with different kernel sizes.

Chain-structured search spaces are conceptually simple, making them easy to design and implement. 
They also often contain strong architectures that can be found relatively quickly.
Their main downside is that, due to the simple architecture topology, there is a comparatively lower chance of discovering a truly novel architecture.

\subsection{Cell-based Search Spaces} \label{subsec:cell}

The cell-based search space is perhaps the most popular type of search space in NAS. 
It is inspired by the fact that state-of-the-art human-designed CNNs often consist of repeated patterns, for example, residual blocks in ResNets \citep{zoph2018learning}. 
Thus, instead of searching for the entire network architecture from scratch, \citet{zoph2018learning} proposed to only search over relatively small \emph{cells}, and stack the cells several times in sequence to form the overall architecture. 
Formally, the searchable cells make up the \emph{micro} structure of the search space, while the outer skeleton (the \emph{macro} structure) is fixed.

\begin{figure}
    \centering
    \includegraphics[trim=1cm 3cm 1cm  1cm, clip, width=1.0\linewidth]{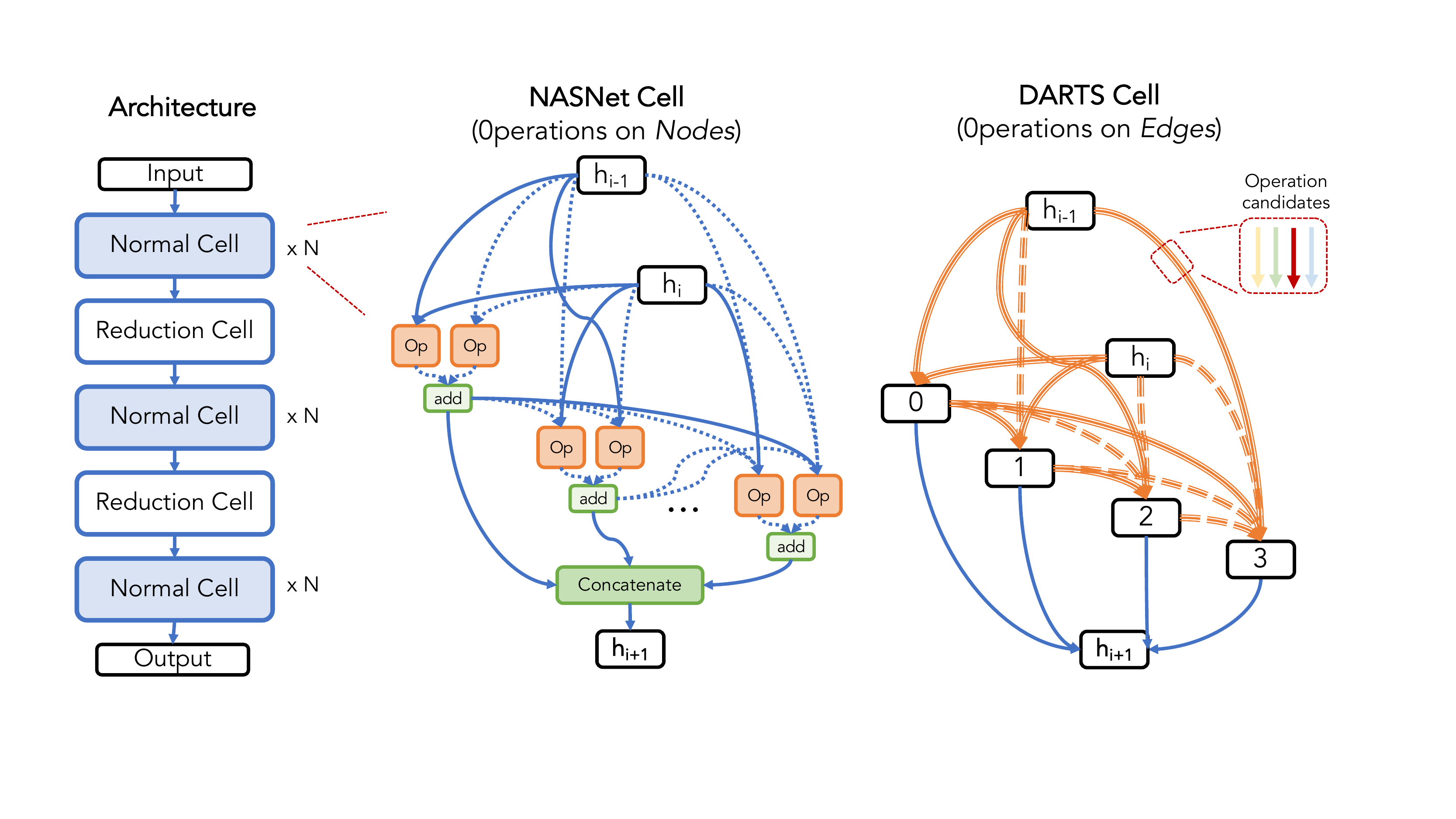}
    \vspace{-2mm}
    \caption{Illustration of cell-based search spaces. 
    The outer skeleton across cells (left) is fixed, while the cells are searchable.
    NASNet assigns operations to nodes (middle) while DARTS assigns operations to edges (right).
    }
    \vspace{-2mm}
    \label{fig:cell_ss}
\end{figure}

The first modern cell-based search space, NASNet, was proposed by \citet{zoph2018learning}. 
It comprises of two types of cells: the \emph{normal} cell and the \emph{reduction} cell. 
Both types have the same structure, but the initial operations in the reduction cell have a stride of two to halve the input spatial resolution. 
Each NASNet cell can be represented as a DAG with seventeen non-input nodes (see Figure \ref{fig:cell_ss} (middle)). 
The nodes are arranged in triples of two operation nodes (such as convolution and pooling operations) and a combination node (such as addition or concatenation). 
The final NASNet architecture is formed by stacking multiple normal and reduction cells in sequence (see Figure \ref{fig:cell_ss} (left)).
Overall, there are $10^{35}$ unique architectures in the NASNet search space.

Since the NASNet search space, many other cell search spaces have been proposed, all of which share a high-level similarity to NASNet, with the main differences being the fixed macro structure, the layout and constraints in the cells, and the choices of operations within the cells.
Two of the most popular cell-based search spaces are NAS-Bench-101 \citep{nasbench} and the DARTS search space \citep{darts}.
NAS-Bench-101 is the first tabular benchmark for NAS (discussed in Section \ref{sec:benchmarks}), and its cells consist of seven nodes, each with three choices of operations; it contains 423\,624 unique architectures. 
The DARTS search space differs more fundamentally: while it also has two searchable cells, the DARTS cells have operation choices on the \emph{edges} of the graph rather than on the nodes. 
In the DARTS cell, the nodes represent latent representations and the edges are operations, whereas in the NASNet cell, the latent representations are on the edges and the nodes are operations.
%
The DARTS cells (see Figure \ref{fig:cell_ss} (right)) contain eight edges, each of which have eight choices of operations. Overall, the DARTS space contains a total of $10^{18}$ unique architectures.

Besides image classification, similar cell designs have also been adopted for language models. 
For example, NAS-Bench-ASR \citep{nasbenchasr} provides a search space of convolutional speech model cells for automatic speech recognition, and there are several LSTM-based search spaces \citep{pham2018efficient,nasbenchnlp,darts}.

The cell-based design significantly reduces the complexity of search spaces, while often resulting in a high-performing final architecture.
This has led to the cell-based search spaces being the most popular type of search space in recent years.
Furthermore,  by detaching the depth of an architecture from the search, the cell-based structure is transferable: the optimal cells learned on a small dataset (e.g., CIFAR-10) typically transfer well to a large dataset (e.g., ImageNet) by increasing the number of cells and filters in the overall architecture  \citep{zoph2018learning, darts}.

Despite their popularity, cell-based search spaces face some criticisms. 
First, while the DARTS search space contains a seemingly large number of $10^{18}$ architectures, the variance in the performance of DARTS architectures is rather small \citep{yang2019evaluation, waniclr2022}.
This small variance may contribute to the fact that sophisticated search strategies can only give marginal gains over the average performance of randomly sampled architectures \citep{yang2019evaluation}. 
Moreover, there are many ad-hoc design choices and fixed hyperparameters that come with cell-based search spaces whose impact is unclear \citep{waniclr2022}, such as the separation of normal and reduction cells, number of nodes, and set of operations.
Finally, although limiting the search to a cell significantly reduces the search complexity, this practice reduces the expressiveness of the NAS search space, making it difficult to find highly novel architectures with cell search spaces. 
In light of this, some recent work advocates for searching for macro connections among cells in addition to the micro cell structure. We discuss this in more detail in the next section.

\subsection{Hierarchical Search Spaces} \label{subsubsec:hierarchical}

Up to this point, all search spaces described have had a \emph{flat} representation, 
in which an architecture is built by defining its hyperparameters, topology, and operation primitives in a single design level. 
Specifically, only one level of topology is searched, whether at the cell level or architecture level.
On the other hand, \emph{hierarchical} search spaces involve designing motifs at different levels, where each higher-level motif is often represented as a DAG of lower-level motifs \citep{liu2018hierarchical,ru2020nago, chrostoforidis2021novel}.

A simple class of hierarchical search spaces has two searchable levels by adding macro-level architecture hyperparameters to cell or chain-structured search spaces.
For example, the MnasNet search space \citep{mnasnet} uses MobileNetV2 as the backbone.
\citet{liu2019auto} designed a two-level search space for semantic image segmentation, and follow-up work extended it to image denoising \citep{zhang2020memory} and stereo matching \citep{kumari2016survey}.
Finally, \citet{chen2021glit} propose a two-level transformer-based search space for vision tasks inspired by ViT \citep{dosovitskiy2020image} and DeiT \citep{touvron2021training}.
The search space consists of a number of sequential blocks which can be a combination of local (convolution) or global (self-attention) layers.

\begin{wrapfigure}{t}{4.0in} 
\begin{minipage}{4.0in}
\centering
\resizebox{4.0in}{!}{\includegraphics{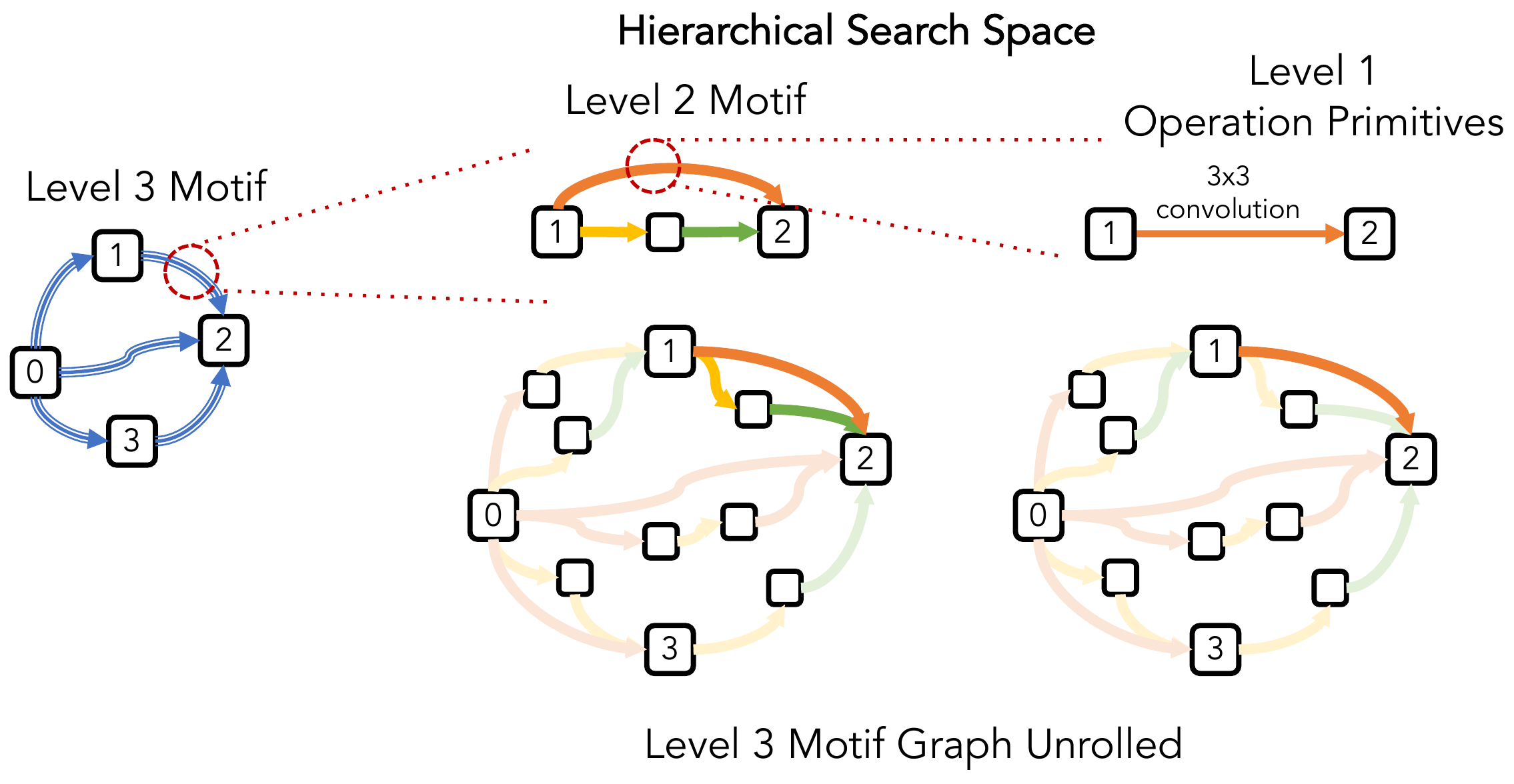}}
    \vspace{-3mm}
    \caption{Illustration of hierarchical representation proposed in \citet{liu2018hierarchical}. Level 1 of the hierarchy consists of choices of operation primitives. Level 2 consists of selecting the topology across small sets of operation primitives.
    Level 3 consists of selecting the topology across the constructions from level 2.
    }
\vspace{-3mm}
\label{fig:hier_rep}
\end{minipage}
\end{wrapfigure}

Beyond two levels, \citet{liu2018hierarchical} and \citet{wu2021trilevel} propose hierarchies of three levels.
\citet{liu2018hierarchical} propose a three-level hierarachy, where each level is a graph made up of components from the previous level (see Figure \ref{fig:hier_rep}).
\citet{wu2021trilevel} propose a different three-level hierarchy, consisting of kernel hyperparameters, cell-based hyperparameters, and macro hyperparameters.
The former design is extended beyond three levels in two follow-up works:
\citet{ru2020nago} proposed a hierarchical design of four levels, controlled by a set of hyperparameters corresponding to a random graph generator, and
\citet{chrostoforidis2021novel} introduced a recursive building process to permit a varying number of hierarchical levels as well as a flexible topology among top-level motifs.

There are multiple benefits to using hierarchical search spaces. 
First, hierarchical search spaces tend to be more expressive.
Most chain-structured, cell-based, and macro search spaces can be seen as a hierarchical search space with a single searchable level, but having two or more levels allows us to search over more diverse and complex architecture designs.
Furthermore, a hierarchical representation of a large architecture is an effective way to reduce the search complexity, which can lead to better search efficiency \citep{liu2018hierarchical, ru2020nago, chrostoforidis2021novel}.
On the other hand, hierarchical search spaces can be more challenging to implement and search through.

\subsection{Architecture Encodings} \label{subsec:encodings}

Throughout this section, we have discussed a wide variety of NAS search spaces. 
As a segue into the next two sections focusing on search strategies, we note that many NAS algorithms and subroutines need to have a succinct representation of each architecture, or \emph{encoding}, in order to perform operations such as mutating an architecture, quantifying the similarity between two architectures, or predicting the test performance of an architecture.
This makes architecture encodings important for several areas of NAS, including discrete NAS algorithms (Section \ref{sec:discrete})
and performance prediction (Section \ref{subsec:prediction}).

In most search spaces, the architecture can be represented compactly as a directed acyclic graph (DAG), where each node or edge represents an operation. 
For example, architectures in cell-based search spaces and chain-structured search spaces can be represented in this way. 
However, hierarchical search spaces cannot be represented fully using a DAG, and often need a conditionally-structured encoding, where the number of levels of conditional hyperparameters correspond to the number of levels of the hierarchy.

For cell-based search spaces, one of the most commonly-used encodings is the adjacency matrix along with a list of operations, of the searchable cell(s) \citep{nasbench, zoph2017neural}.
In order to have better generalizablility, \citet{ning2020generic} proposed a graph-based encoding scheme  and \citet{bananas} proposed a path-based encoding scheme, both of which model the flow of propagating information in the network.
Finally, another type of encoding for all search spaces is a \emph{learned} encoding using unsupervised pre-training. 
In this technique, before we run NAS, we use a set of untrained architectures to learn an architecture encoding, for example, by using an autoencoder \citep{yan2020does,dvae,li2020neural,lukasik2022learning,lukasik2021smooth} or a transformer \citep{yan2021cate}.

When choosing an architecture encoding, scalability and generalizability are important traits.
Recent work has shown that different NAS subroutines, such as sampling a random architecture, perturbing an architecture, or training a surrogate model, may each perform best with different encodings \citep{white2020study}.
Furthermore, even small changes to the architecture encoding scheme can have significant effects on the performance of NAS \citep{nasbench,white2020study}.

\section{Black-Box Optimization Techniques} \label{sec:discrete}

Now that we have covered search spaces, we move to perhaps the most widely-studied component of NAS: the \emph{search strategy}. 
This is what we run to find an optimal architecture from the search space. 
Search strategies generally fall into two categories: black-box optimization techniques and one-shot techniques.
However, some methods that we discuss include characteristics of both, or neither, of these categories.
We first discuss black-box optimization techniques in this section, followed by one-shot techniques in Section \ref{sec:oneshot}.

For black-box optimization, we discuss baselines (Section \ref{subsec:baselines}), reinforcement learning (Section \ref{subsec:reinforcement_learning}), evolution (Section \ref{subsec:evolution}), Bayesian optimization (Section \ref{subsec:bayesian_optimization}), and Monte-Carlo tree search (Section \ref{subsec:mcts}).
Black-box optimization techniques are widely used and studied today, due to their strong performance and ease of use.
In general, black-box optimization techniques tend to use more computational resources than one-shot techniques, due to training many architectures independently (without sharing weights across architectures like one-shot techniques).
However, they also have many advantages over one-shot techniques, such as robustness (and the lack of catastrophic failure modes), simpler optimization of non-differentiable objectives, simpler parallelism, joint optimization with other hyperparameters, and easier adaptation to, e.g., new problems, datasets or search spaces.
They are also often conceptually simpler, making them easier to implement and use.


\subsection{Baselines} \label{subsec:baselines}

One of the simplest possible baselines for NAS is \emph{random search}: architectures are selected randomly from the search space and then fully trained. In the end, the architecture with the best validation accuracy is outputted.
Despite its na\"ivet\'e, multiple papers have shown that random search performs surprisingly well \citep{sciuto2019evaluating, randomnas, yang2019evaluation, dpc}. 
This is especially true for highly engineered search spaces with a high fraction of strong architectures, since random search with a budget of $k$ evaluations will, in expectation, find architectures in the top $100/k\%$ of the search space.
However, other works show that random search does not perform well on large, diverse search spaces \citep{Bender_2020_CVPR, real2020automl}.
Still, random search is highly recommended as a baseline comparison for new NAS algorithms \citep{yang2019evaluation, lindauer2019best}, and can be made highly competitive by incorporating weight sharing \citep{randomnas}, zero-cost proxies \citep{abdelfattah2021zerocost}, or learning curve extrapolation \citep{nasbenchx11}.
Multiple papers \citep{sciuto2019evaluating, yang2019evaluation} have also proposed a related, simpler baseline: \emph{random sampling}, the average performance of architectures across the entire search space.

In addition to random search, recent papers showed that local search is a strong baseline for NAS on both small \citep{ottelander2020local,white2021exploring} and large \citep{nasbench301} search spaces.
This is true even for the simplest form of local search: iteratively train and evaluate all of the neighbors of the best architecture found so far, where the neighborhood is typically defined as all architectures which differ by one operation or edge.
Local search can be sped up substantially by using network morphisms to warm-start the optimization of neighboring architectures \citep{elsken2017simple}.

\subsection{Reinforcement Learning} \label{subsec:reinforcement_learning}

Reinforcement learning (RL) was very prominent in the early days of modern NAS. 
Notably, the seminal work by \citet{zoph2017neural} used RL on 800 GPUs for two weeks to obtain competitive performance on CIFAR-10 and Penn Treebank; this finding received substantial media attention and started the modern resurgence of NAS.
This was followed up by several more reinforcement learning approaches \citep{pham2018efficient, zoph2018learning}.

\begin{algorithm}[t]
	\begin{algorithmic}
        \STATE \textbf{Input:} Search space $\archss{}$, number of iterations $T$.
        \STATE Randomly initialize weights $\theta$ of the controller architecture.\\
	\FOR{$t=1, \dots, T$}
		\STATE
                Train architecture $a \sim \pi(a; \theta)$, randomly sampled from the controller policy $\pi(a; \theta)$.\\
            \STATE
                Update controller parameters $\theta$ by performing a gradient update $\nabla_\theta E_{a \sim \pi(a; \theta)}[\lval(a)]$.
	\ENDFOR   
      \STATE \textbf{Output:} Architecture selected from the trained policy $\pi(a; \theta^*)$
	\end{algorithmic}
	\caption{General Reinforcement Learning NAS Algorithm}
	\label{alg:nasrl1}
\end{algorithm}

Most reinforcement learning approaches model the architectures as a sequence of actions generated by a controller \citep{zoph2017neural, baker2016designing}. 
The validation accuracy of the sampled architectures after training is used as a reward signal to update the controller in order to maximize its expected value. See Algorithm \ref{alg:nasrl1}.
The controller is usually a recurrent neural network (RNN) \citep{zoph2017neural,zoph2018learning} that outputs a sequence of components corresponding to an architecture.
After each outputted architecture is trained and evaluated, the RNN parameters are updated to maximize the expected validation accuracy of outputted architectures, using REINFORCE \citep{williams92,zoph2017neural} or proximal policy optimization \citep{Schulman2017ProximalPO,zoph2018learning}.
ENAS \citep{pham2018efficient} follows a similar strategy but speeds up the reward estimation using weight sharing; we will discuss this in detail in Section \ref{sec:oneshot}.

More recently, RL has not been used prominently for NAS, since it has been shown to be outperformed in head-to-head comparisons by evolutionary methods \citep{real2019regularized} and Bayesian optimization \citep{nasbench}, which we will discuss next.


\subsection{Evolutionary and Genetic Algorithms}\label{subsec:evolution}

Decades before the recent NAS resurgence, one of the first works in NAS used an evolutionary algorithm \citep{miller1989designing}.
In other early works, it was common to use evolutionary algorithms to simultaneously optimize the neural architecture and its weights 
\citep{angeline1994evolutionary, stanley2002evolving, floreano2008neuroevolution, stanley2009hypercube}.
Today, evolutionary algorithms are still popular for the optimization of architectures due to their flexibility, conceptual simplicity, and competitive results \citep{real2019regularized}, but the weight optimization is typically left to standard SGD-based approaches.


Evolutionary NAS algorithms work by iteratively updating a population of architectures.
In each step, one or more ``parent'' architectures in the population are sampled (typically based on the validation accuracy of the architectures), combined and mutated to create new ``children'' architectures. 
These architectures are then trained and added to the population, replacing individuals in the population with worse performance.
See Algorithm \ref{alg:general_ea}.

There are many other ways in which evolutionary algorithms differ, including sampling the initial population, selecting the parents, and generating the children.
For selecting the initial population, approaches include using trivial architectures \citep{Real17}, randomly sampling architectures from the search space \citep{real2019regularized, sun2019evolving}, or using hand-picked high-performing architectures \citep{fujino2017deep}.

\begin{algorithm}[t]
\begin{algorithmic}
        \STATE \textbf{Input:} Search space $\archss{}$, number of iterations $T$.  
        \STATE Randomly sample and train a population of architectures from the search space $\archss{}$. 
	\FOR{$t=1, \dots, T$}
	\STATE Sample (based on accuracy) a set of parent architectures from the population.
	\STATE Mutate the parent architectures to generate children architectures, and train them.
        \STATE Add the children to the population, and kill off the architectures that are the oldest (or have the lowest accuracy) among the current population.
	\ENDFOR 
      \STATE \textbf{Output:} Architecture from the population with the highest validation accuracy.
\end{algorithmic}
\caption{General Evolutionary NAS Algorithm}\label{alg:general_ea}
\end{algorithm}

Selecting parents from the population makes up one of the core components of the evolutionary algorithm.
Perhaps the most popular method to sample parents is tournament selection \citep{goldberg1991comparative,Real17, real2019regularized, sun2019evolving, sun2020automatically, almalaq2018evolutionary}, which selects the best architecture(s) out of a randomly sampled population.
Other common approaches include random sampling weighted by fitness \citep{xie2017genetic, gibb2018genetic, song2020efficient, loni2020deepmaker}, or choosing the current best architecture(s) as 
parents \citep{elsken2017simple, suganuma2018exploiting, suganuma2017genetic}.
These methods trade off exploration vs.\ exploiting the best region found so far.
One particularly successful evolutionary algorithm is \emph{regularized evolution} by \citet{real2019regularized}. 
This is a fairly standard evolutionary method, with the novelty of dropping the architecture in each step that has been in the population for longest, even if it has the highest performance.
This method outperformed random search and RL in a head-to-head comparison and achieved state-of-the-art performance on ImageNet at the time of its release \citep{real2019regularized}.

\subsection{Bayesian Optimization} \label{subsec:bayesian_optimization}


Bayesian optimization (BO, see, e.g.\ \citet{frazier2018tutorial} or \citet{garnett_bayesoptbook_2023}) is a powerful method for optimizing expensive functions, and it has seen significant success within NAS.
There are two key components to BO: \emph{(1)} building a probabilistic surrogate to model the unknown objective based on past observations, and \emph{(2)} defining an acquisition function to balance the exploration and exploitation during the search. 
BO is an iterative algorithm which works by selecting the architecture that maximizes the acquisition function (computed using the surrogate), training this architecture, and retraining the surrogate using this new architecture to start the next iteration.
See Algorithm \ref{alg: general_seq_BO}. 

Initial BO-based NAS techniques developed custom distance metrics among architectures, for example, with a specialized architecture kernel \citep{swersky2014raiders}, an optimal transport-inspired distance function \citep{nasbot}, or a tree-Wasserstein distance function \citep{nguyen2021optimal}, allowing a typical Gaussian process (GP) based surrogate with BO.
However, using a standard GP surrogate often does not perform well for NAS, as search spaces are typically high-dimensional, non-continuous, and graph-like.
To overcome this, one line of work first encodes the architectures, using encodings discussed in Section \ref{subsec:encodings}, and then trains a model, such as a tree-Parzen estimator \citep{tpe, bohb}, random forest \citep{hutter11, nasbench}, or neural network \citep{bananas, springenberg2016bayesian}.
Another line of work projects architecture information into a low-dimensional continuous latent space on which conventional BO can be applied effectively \citep{ru2020nago,wan2022approximate}. 
Another class of surrogate models use graph neural networks \citep{ma2019deep,shi2020bonas,nasbowl} or a graph-based kernel \citep{nasbowl} to naturally handle the graph representation of architectures without the need for an explicit encoding.

\begin{algorithm}[t]
\begin{algorithmic}
        \STATE \textbf{Input:} Search space $\archss{}$, number of iterations $T$, acquisition function $\phi$.
        \STATE Randomly sample and train a population of architectures from the search space $\archss{}$. 	
        \FOR{$t=1, \dots, T$}
        \STATE Train a surrogate model based on the current population.
	\STATE Select architecture $a_t$ by maximizing $\phi\left(a\right),$ based on the surrogate model.
	\STATE Train architecture $a_{t}$ and add it to the current population.
	\ENDFOR
      \STATE \textbf{Output:} Architecture from the population with the highest validation accuracy. 
\end{algorithmic}
\caption{General Bayesian Optimization NAS Algorithm}\label{alg: general_seq_BO}
\end{algorithm}

The acquisition function, which trades off exploration and exploitation during the search, is another important design component for BO. 
There are various types of acquisition functions used in NAS, such as expected improvement \citep{movckus1975bayesian,jones1998efficient}, upper confidence bound \citep{cox1992statistical,srinivas2010gaussian} and information-theoretic ones \citep{hennig2012entropy, hernandez2014predictive, wang2017max, hvarfner2022joint}.
In NAS, optimizing the acquisition function in each round of BO is challenging due to the non-continuous search spaces, and furthermore, exhaustively evaluating acquisition function values on all possible architectures is computationally non-viable. 
The most common method for optimizing the acquisition function in NAS is by randomly mutating a small pool of the best architectures queried so far, and of the mutated architectures, selecting the one(s) with the highest acquisition function value \citep{nasbot, ma2019deep, bananas, shi2020bonas, nasbowl,schneider2021mutation}. 
Other methods for optimizing the acqusition function include local search, evolutionary search, and random search \citep{nasbench, nasbowl, shi2020bonas}.

\subsection{Monte Carlo Tree Search} \label{subsec:mcts}

Another class of NAS methods is based on Monte Carlo Tree Search (MCTS). 
MCTS is the key backbone search algorithm used in AlphaGO \citep{silver2016mastering} and AlphaZero \citep{silver2017mastering}, which achieve super-human performance in Go and chess, respectively.
MCTS finds optimal decisions by recursively sampling new decisions (e.g., making a move in chess, or selecting an operation for an architecture in NAS), running stochastic \emph{rollouts} to obtain the reward (such as winning a chess game, or discovering a high-performing architecture) and then backpropagating to update the weight of the initial decision.
Across iterations, the algorithm builds a decision tree to bias the search towards more promising regions by balancing exploration and exploitation in decision making \citep{browne2012survey}.

MCTS was first applied to NAS by \citet{negrinho2017deeparchitect} who represented the search space and its hyperparameters using a modular language.
This results in a tree-structured, extensible search space, contrary to the fixed search spaces of prior work. 
\citet{wistuba2017finding} introduced a similar method but with two different UCT (Upper Confidence bounds applied to Trees) algorithms. 
MCTS was first adapted to cell-based search spaces by using a state-action representation \citep{alphax}. The authors also improved sample efficiency by using a neural network to estimate the accuracy of sampled architectures, thus enabling a higher number of rollouts.
This was followed up by adding further efficiency in pruning the tree by learning partitionings \citep{wang2020neural}, and by application to multi-objective NAS \citep{zhao2021multi}.

\section{One-Shot Techniques} \label{sec:oneshot}

Throughout Section \ref{sec:discrete}, we have seen that the predominant methodology in the early stages of NAS research was to iteratively sample architectures from the search space, train them, and use their performance to guide the search.
The main drawback of these methods, when applied without speedup techniques, is their immense computational cost, sometimes on the order of thousands of GPU days \citep{zoph2017neural,real2019regularized} due to the need to train thousands of architectures \emph{independently} and \emph{from scratch}.\footnote{
On the other hand, recent developments in performance estimation and speed-up techniques (Section \ref{sec:speedup}) have significantly improved the computational overhead of methods that use black-box optimization as a base, making these methods affordable for many applications and users.
}

As an alternative, \emph{one-shot} techniques were introduced to avoid training each architecture from scratch, thus circumventing the associated computational burden. 
As of 2022, they are currently one of the most popular techniques in NAS research.
Rather than training each architecture from scratch, one-shot approaches implicitly train all architectures in the search space via a single (``one-shot'') training of a \emph{hypernetwork} or \emph{supernetwork}.

\begin{figure}[t]
\centering
\includegraphics[trim=0cm 5cm 0cm  5cm, clip, width=1.0\linewidth]{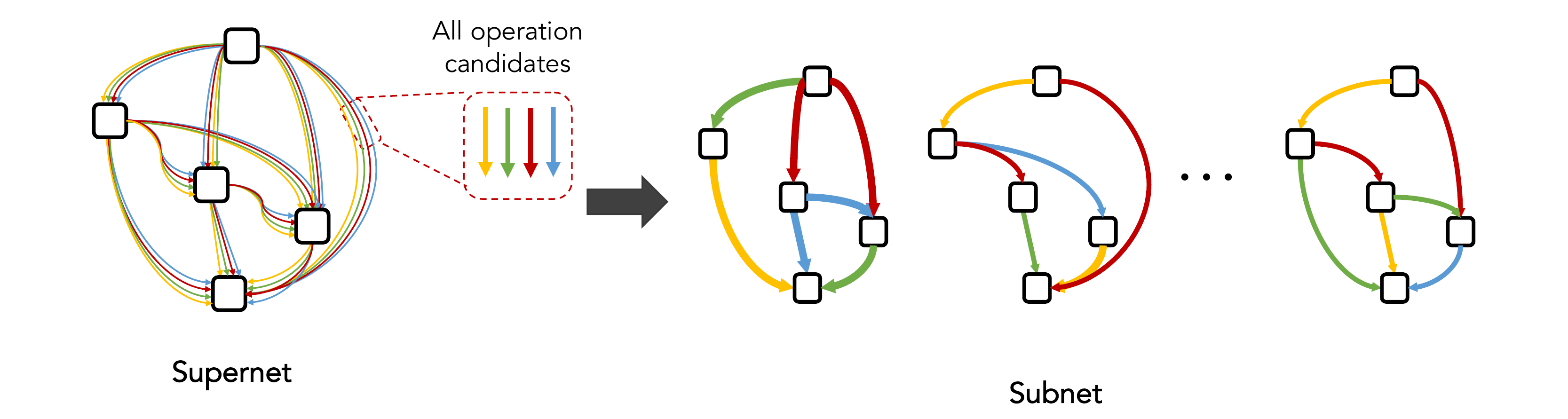}
    \vspace{-4mm}
\caption{A supernet comprises all possible architectures in the search space. Each architecture is a subnetwork (subgraph) in the supernet.
}
\label{fig:supernet_illustation}
\end{figure}

A \emph{hypernetwork} is a neural network which generates the weights of \emph{other} neural networks \citep{schmidhuber1992learning}, while a \emph{supernetwork} (often used synonymously with ``one-shot model'' in the literature) is an over-parameterized architecture that contains all possible architectures in the search space as subnetworks (see Figure \ref{fig:supernet_illustation}).
The idea of a supernetwork was introduced by \citet{SaxenaV16} and was popularized in 2018 by works such as
\citet{bender2018understanding}, \citet{pham2018efficient}, and \citet{darts}.

Once a supernet is trained, each architecture from the search space can be evaluated by inheriting its weights from the corresponding subnet within the supernet.
The reason for the scalability and efficiency of supernets is that a linear increase in the number of candidate operations only causes a linear increase in computational costs for training, but the number of subnets in the supernet increases exponentially. Therefore, supernets allow us to train an exponential number of architectures for a linear compute cost.

A key assumption made in one-shot approaches is that when using the one-shot model to evaluate architectures, the ranking of architectures is relatively consistent with the ranking one would obtain from training them independently. 
The extent to which this assumption holds true has been substantially debated, with work showing evidence for \citep{yu2020train, li2020geometry, pham2018efficient} and against \citep{sciuto2019evaluating, nasbench1shot1, zhang2020deeper, pourchot2020share} the claim across various settings.
The validity of the assumption is dependent on the search space design, the techniques used to train the one-shot model, and the dataset itself, and it is hard to predict to what degree the assumption will hold in a particular case \citep{sciuto2019evaluating,zhang2020deeper}.

\begin{figure}[t]
\centering
\includegraphics[width=1.0\linewidth]{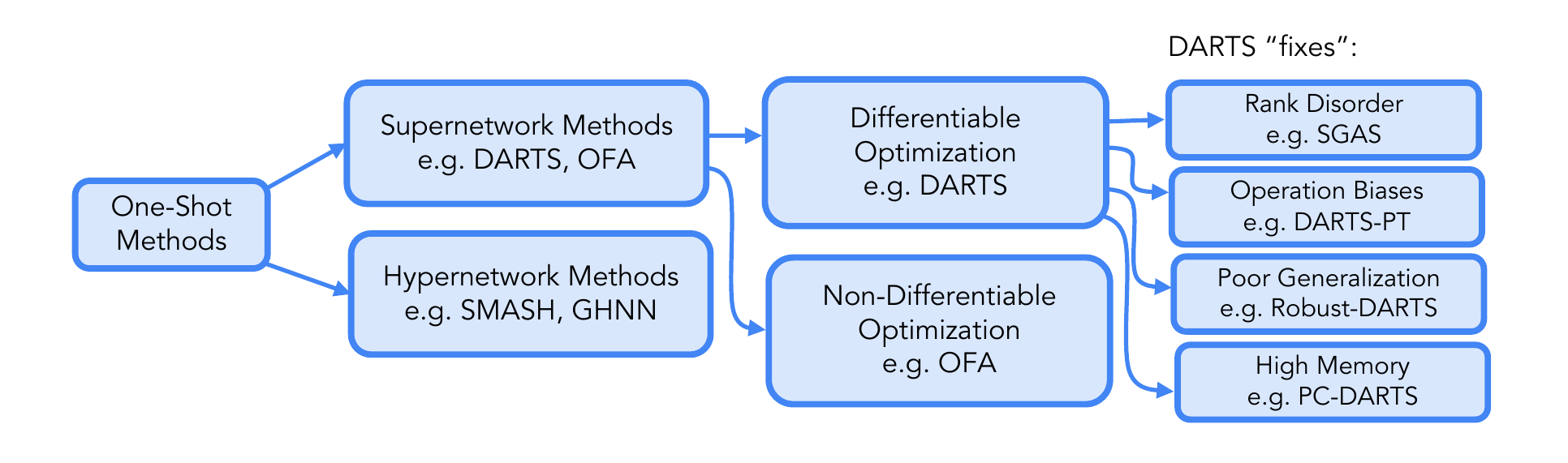}
    \vspace{-4mm}
\caption{A taxonomy of the predominant one-shot families.
A \emph{hypernetwork} is a neural net which generates the weights of other neural nets.
A \emph{supernetwork} is an over-parameterized neural net that contains the set of neural nets from the search space as subnetworks, and it can be used with differentiable optimization (including DARTS and follow-ups), or non-differentiable optimization.
}
\label{fig:oneshot_taxonomy}
\end{figure}

While the supernet allows quick evaluation of all architectures, we must still decide on a search strategy, which can be as simple as running a black-box optimization algorithm while the supernet is training (such as in \citet{pham2018efficient}) or after the supernet is trained (such as in \citet{bender2018understanding}).
We discuss these families of techniques in Section \ref{subsec:non-darts}.
A popular line of work uses gradient descent to optimize the architecture hyperparameters in tandem with training the supernet (such as DARTS \citep{darts} and numerous subsequent methods).
We discuss this family of techniques in Section \ref{subsec:darts}.
Finally, in Section \ref{subsec:hypernetworks}, we discuss hypernetworks.
Figure \ref{fig:oneshot_taxonomy} provides a taxonomy of one-shot families.

\subsection{Non-Differentiable Supernet-Based Methods} \label{subsec:non-darts}
We start by describing supernet-based methods which do not make use of differentiable optimization.
Some methods in this family decouple the supernet training and architecture search: first train a supernet, and then run a black-box optimization algorithm to search for the best architecture. 
Other methods train a supernet while simultaneously running a non-differentiable search algorithm, such as reinforcement learning, to select subnetworks.

\citet{bender2018understanding}, \citet{randomnas}, and \citet{guo2020single} propose simple methods to train the supernet and then use a black-box optimization algorithm to extract the best architecture from it.
\citet{bender2018understanding} construct the supernet by creating a separate node corresponding to an operation, in every place where there is a choice of operation; they then train the supernet as if it were a standard neural net, with one exception: nodes are randomly dropped during training, with the level of dropout increasing linearly throughout training.
In follow-up work, \citet{randomnas} and \citet{guo2020single} take this idea a step further: in each training step, they randomly sample one architecture and only update the weights of the supernet corresponding to that architecture.
These techniques better mimic what is happening at evaluation time: only a subnetwork is evaluated rather than the entire supernet.
Furthermore, these procedures use significantly less memory than training all the weights of a supernet.
Each method concludes by using the trained supernet to quickly evaluate architectures when conducting random search \citep{bender2018understanding,randomnas} or evolutionary search \citep{guo2020single}. The architecture identified in the end is then trained from scratch.

As will be discussed in Section \ref{subsec:hardware}, deploying neural nets in practice often comes with constraints on latency or memory.
While the supernets considered thus far tend to only contain architectures of approximately the same size, \citet{ofa} propose a supernet containing subnetworks of various sizes. This \emph{Once-for-all (OFA)} approach uses a progressive shrinking strategy which starts by sampling the largest subnetworks, and then moving to smaller subnetworks, in order to minimize the co-adaptation among subnetworks and effectively train networks of different sizes ``once for all''.
In a subsequent \emph{search phase}, architectures are selected based on different constraints on latency and memory.
While \citet{ofa} uses random search for this search phase, \citet{guo2020single} proposed to improve this approach further by using evolutionary search in the search phase.

One of the earliest supernet-based approaches is ENAS (Efficient Neural Architecture Search) \citep{pham2018efficient}, which trains the supernet while running a search algorithm in tandem.
Specifically, the search strategy is similar to the RL controller-based approach from \citet{zoph2017neural} (described in Section \ref{subsec:reinforcement_learning}) but estimates the performance of each architecture using a supernet. 
The training procedure alternates between selecting an architecture, evaluating it, and updating the weights of the supernet, and updating the weights of the controller by sampling several architectures to estimate the reward of REINFORCE.
%
While this approach searches for an architecture in tandem with training the supernet, it uses a separate controller network to guide the search. In the next section, we discuss methods which conduct the search via gradient descent using only the supernet.


\begin{algorithm}[t]
\begin{algorithmic}
\STATE \textbf{Input:} Search space $\archss{}$, number of iterations $T$, hyperparameter $\xi$.
    \STATE Randomly initialize a one-shot model based on $\archss{}$ with weights $w$ and architecture hyperparameters $\alpha$.
    \FOR{$t=1, \dots, T$}
\STATE
    Perform a gradient update on the architecture weights $\alpha$ according to Equation \ref{eq:darts}.\\
    Perform a gradient update on $w$ according to $\nabla_w \mathcal{L}_{train}(w, \alpha).$
\ENDFOR
\STATE \textbf{Output:} Derive the final architecture by taking the argmax of $\alpha$, across all operation choices, and then retrain this architecture from scratch. 
\end{algorithmic}
\caption{DARTS - Differentiable Architecture Search}
\label{alg:darts}
\end{algorithm}  

\subsection{Differentiable Supernet-Based Methods} \label{subsec:darts}
In this section, we review supernet-based NAS methods that employ differentiable optimization techniques. 
We first describe the seminal DARTS (Differentiable Architecture Search) approach by \citet{darts}, and then we move to various follow-up works and other differentiable approaches.

The DARTS approach uses a continuous relaxation of the discrete architecture search space, which enables the use of gradient descent in order to find a high-performing local optimum significantly faster than black-box optimization methods.
It can be applied to any DAG-based search space which has different choices of operations on each edge by using a ``zero'' operation to simulate the absence of an edge.

\begin{figure}[t]
\centering
\includegraphics[width=1.0\linewidth]{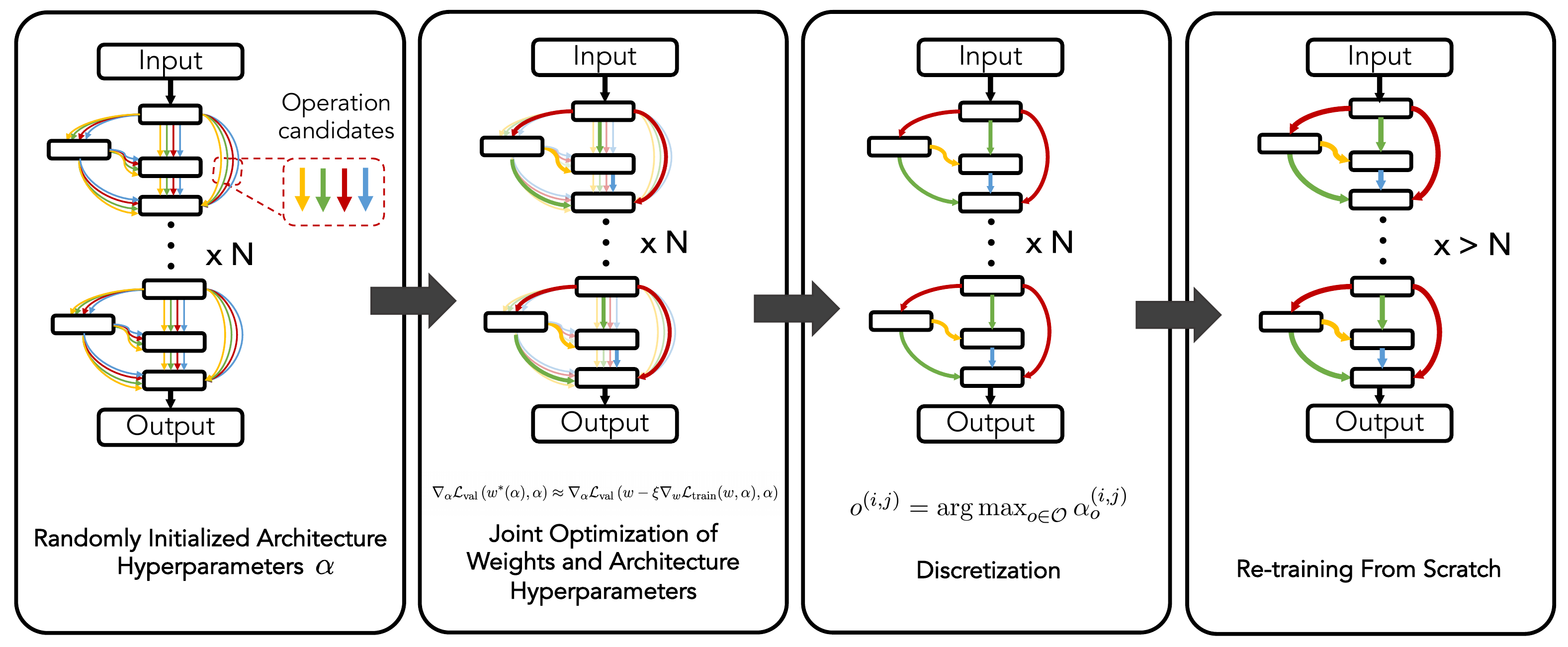}
\caption{Differentiable one-shot NAS algorithms have four main steps:
randomly initializing the architecture hyperparameters, optimizing the architecture hyperparameters and weights via alternating gradient descent, discretizing the optimized architecture hyperparameters, and re-training the resulting subnetwork from scratch.
}
\label{fig:darts-diagram}
\end{figure}

At the start, each edge $(i,j)$ in the DARTS search space consists of multiple possible candidate operations $o$, each of which are associated with a continuous hyperparameter $\alpha_o^{(i,j)}\in [0,1]$.
While the supernet is training, edge $(i,j)$ consists of a mix of all candidate operations, weighted by each $\alpha_o^{(i,j)}$.
The architecture hyperparameters $\alpha$ are optimized jointly with the supernet model weights $w$ via alternating gradient descent. 
In particular, in order to update the architecture weights $\alpha$ via gradient descent, DARTS makes use of the following approximation:
\begin{equation} \label{eq:darts}
\nabla_\alpha \lval\left(w^*(\alpha),\alpha\right) 
\approx \nabla_\alpha \lval\left(w - \xi\nabla_w\ltrain(w,\alpha),\alpha\right),
\end{equation}
where $\ltrain$ denotes the training loss, $\lval$ denotes the validation loss, $\xi$ is the learning rate, and $w^*(\alpha)$ denotes the weights that minimize the training loss of the architecture corresponding to $\alpha$.
In other words, in order to avoid the expensive inner optimization, $w^*(\alpha)$ is approximated by a single step of gradient descent ($w-\xi\nabla_w\ltrain(w,\alpha)$).
This is similar to MAML \citep{finn_maml} and other works \citep{luketina2016scalable,metz2017unrolled}.
Although this strategy is not guaranteed to converge, \citet{darts} showed that it works well in practice with a suitable choice of $\xi$.
%
%
After the training phase, DARTS obtains a discrete architecture by selecting the operation with the maximum value of $\alpha$ on each edge (the discretization step) and then re-trains it from scratch.
Figure \ref{fig:darts-diagram} provides an illustration of DARTS.

DARTS gained significant attention in the AutoML community due to its simplicity, its novelty, and the release of easy-to-use code.
Furthermore, the original technique left room for improvement across various axes.
Consequently, there has been a large body of follow-up work seeking to improve various parts of the DARTS approach. 
In the rest of the section, we cover the main categories of improvements (see Figure \ref{fig:oneshot_taxonomy}).

\subsubsection{Rank Disorder} \label{subsubsec:disorder}

As mentioned at the start of Section \ref{sec:oneshot}, nearly all one-shot methods make a key assumption: the ranking of architectures evaluated with the supernet is relatively consistent with the ranking one would obtain from training them independently; when this assumption is not met, it is known as \emph{rank disorder} \citep{li2020geometry,sciuto2019evaluating}. 
While there is considerable debate both for \citep{yu2020train, li2020geometry, pham2018efficient} and against \citep{sciuto2019evaluating, nasbench1shot1, zhang2020deeper, pourchot2020share} the assumption, many works have attempted to reduce the problem of rank disorder.

Several methods propose to gradually increase the network depth, or to gradually prune the set of operation candidates during training, showing that this causes the weights to better adapt to the most-promising operation choices.
Progressive-DARTS \citep{chen2019progressive} gradually increases the network depth while simultaneously pruning the operations with the smallest weights.
SGAS \citep{li2020sgas} chooses operations throughout the training procedure, based on two criteria: selection certainty (calculated via the entropy of the operation distribution) and selection stability (calculated via the movement of the operation distribution).
Finally, XNAS \citep{nayman2019xnas} makes use of the exponentiated gradient algorithm \citep{kivinen1997exponentiated}, which dynamically prunes inferior operation choices during the search while also allowing the recovery of ``late bloomers'', i.e., operation choices which only become accurate later in the training procedure.

\subsubsection{Operation Biases}

Several works show that differentiable NAS techniques tend to favor skip connections over other operation choices \citep{zela2020understanding,wang2021rethinking,darts+}, which might be caused by the supernet using skip connections to over-compensate for vanishing gradients \citep{chu2020darts}. 
Various methods have been proposed to fix this bias.

DARTS$+$ \citep{darts+} proposes an early stopping method based on the stability of the ranking of the architecture weights, while
DARTS$-$ \citep{chu2020darts} separates the skip connection weights from other operation weights via auxiliary edges.
FairDARTS \citep{chu2020fair} sets \emph{all} operation weights independent
of all others, and then pushes these architecture weights toward zero or one in the loss function.

Taking a different approach, \citet{wang2021rethinking} show that it is okay for skip connections to have higher weights, as long as we do not select the final architecture based on these weights. 
Instead, after training the supernet, their algorithm, DARTS-PT, selects each operation whose removal has the largest decrease of accuracy in the supernet.

Rather than fixing the biases among a small hand-picked set of operations, \citet{shen2022efficient} instead use a search space that significantly reduces human bias: they fix a standard convolutional network and search for the kernel sizes and dilations of its operations.
This simple approach is broadly applicable across computer vision, PDE solving, protein folding, and other tasks. 
In order to make one-shot training more efficient, their algorithm, DASH, computes the mixture-of-operations using the Fourier diagonalization of convolution.

\subsubsection{Poor Test Generalization}

Several works seek to improve the generalization performance of DARTS through various means.
\citet{zela2020understanding} and \citet{chen2020stabilizing} show that DARTS often converges to sharp local minima in the loss landscape (high validation loss curvature in the architecture hyperparameter space), which, after running the discretization step, can cause the algorithm to return an architecture with poor test generalization.
Robust-DARTS \citep{zela2020understanding} fixes this issue by making the training more robust through data augmentation, $L_2$ regularization of the inner objective $\mathcal{L}_{train}$, and early stopping.
Similarly, rather than optimizing the training loss, Smooth-DARTS \citep{chen2020stabilizing} optimizes the expected or worst-case training loss over a local neighborhood of the architecture hyperparameters.

Taking a different approach, GAEA \citep{li2020geometry}, XD \citep{roberts2021rethinking}, and StacNAS \citep{guilin2019stacnas} all use a single-level optimization rather than the typical bi-level optimization, by treating the architecture hyperparameters as normal architecture weights, showing this leads to better generalization.
Furthermore, GAEA re-parameterizes the architecture parameters over the simplex and updates them using the exponentiated gradient algorithm (similar to XNAS from Section \ref{subsubsec:disorder}), showing this is better-suited to the underlying geometry of the architecture search space.

Finally, Amended-DARTS \citep{bi2019stabilizing} and iDARTS \citep{zhang2021idarts} both take the approach of deriving more accurate approximations of the gradients of $\alpha$ (Equation \ref{eq:darts}), showing that this leads to a more stable optimization and better generalization.

\subsubsection{High Memory Consumption} \label{subsubsec:memory}

The memory required to train a supernet is much higher than a normal neural net---it scales linearly with the size of the set of candidate operations.
Recall from Section \ref{subsec:non-darts} that multiple works reduced this memory by, in each training step, masking out all operations except for the ones corresponding to one or a few subnetworks.
Various works have proposed techniques to mask out operations for differentiable NAS as well, i.e., while simultaneously optimizing the architecture hyperparameters. 

\citet{proxylessnas} proposed ProxylessNAS, which solves this problem by modifying the BinaryConnect \citep{courbariaux2015binaryconnect} discretization method: in each training step, for each operation choice, all are masked out except one operation that is randomly chosen with probability proportional to its current value of $\alpha$.
\citet{proxylessnas} show that this procedure converges to a single high-performing subnetwork.
GDAS \citep{dong2019searching} and DSNAS \citep{xie2018snas,Hu2020DSNASDN} use a Gumbel-softmax distribution over a one-hot encoding of the operation choices, which is a different way to allow sampling single operations in each training step while maintaining differentiability.

PC-DARTS \citep{pcdarts} proposes a relatively simpler approach: at each training step, and for each edge in the DAG, a subset of channels is sampled and sent through the possible operations, while the remaining channels are directly passed on to the output.
While reducing memory due to training fewer channels, this also acts as a regularizer.
DrNAS \citep{chen2021drnas} also reduces memory consumption by progressively increasing the number of channels that are forwarded to the mixed operations, and progressively pruning operation choices, modeled by a Dirichlet distribution.


\subsection{Hypernetworks} \label{subsec:hypernetworks}

A \emph{hypernetwork} is a neural network which generates the weights of \emph{other} neural networks.
Hypernetworks were first considered by \citet{schmidhuber1992learning,schmidhuber1993self}, and the first modern application was by \citet{hypernets}, who used them to obtain better weights for a fixed LSTM architecture.
Hypernetworks have since been used for a variety of tasks, including HPO \citep{navon2021learning,mackay2018self}, 
calibrating model uncertainty \citep{krueger2017bayesian},
and NAS \citep{brock2018smash, zhang2018graph}.

The first work to use hypernetworks for NAS (and among the first to use a one-shot model for NAS) was SMASH (one-Shot Model Architecture Search through Hypernetworks) \citep{brock2018smash}.
SMASH consists of two phases: first, train a hypernetwork to output weights for any architecture in the search space. 
Next, randomly sample a large set of architectures, generate their weights using the hypernetwork, and output the one with the best validation accuracy. 
The hypernetwork, a convolutional neural net, takes as input an architecture encoding and outputs a set of weights for that architecture, and is trained by randomly sampling an architecture, generating its weights, computing its training error, and then backpropagating through the entire system (including the hypernetwork weights).

Another hypernet-based NAS algorithm is GHN (Graph Hypernetworks) \citep{zhang2018graph}.
The main difference between SMASH and GHN is the architecture encoding and the architecture of the hypernetwork.
Specifically, the GHN hypernetwork is a mix between a graph neural network and a standard hypernetwork. It takes as input the computational graph of an architecture $a$ and uses message-passing operations which are typical in GNNs, to output the weights of $a$. 
The training of the hypernetwork, and the final NAS algorithm, are both the same as in SMASH.



\section{Speedup Techniques} \label{sec:speedup}
In this section, we cover general speedup techniques for NAS algorithms, including performance prediction (Section \ref{subsec:prediction}),
multi-fidelity methods (Section \ref{subsec:multi-fidelity}), meta-learning approaches (Section \ref{subsec:meta-learning}), and weight inheritance (Section \ref{subsec:inheritance}).

\subsection{Performance Prediction} \label{subsec:prediction}

A large body of work has been devoted to predicting the performance of neural networks before they are fully trained. 
Such techniques have the potential to greatly speed up 
the runtime of NAS algorithms, since they remove the need to fully train each architecture under consideration.
These speedup techniques can improve nearly all types of NAS algorithms, 
from black-box optimization \citep{ru2020revisiting, white2021powerful}
to one-shot NAS \citep{xiang2021zero}.
In this section, we discuss the performance prediction techniques themselves, while in Section \ref{subsec:multi-fidelity}, we discuss methods of incorporating them into NAS algorithms.

Formally, given a search space \searchspace and architecture $\archmath\in\searchspacemath$, denote the final validation accuracy obtained with a fixed training pipeline as $f(a)$.
A \emph{performance predictor}  $f'$ is defined as any function which predicts
the accuracy or relative accuracy of architectures, without fully training them.
In other words, evaluating $f'(\archmath)$ takes less time than evaluating \nasobjective{}, and $\{f'(\archmath)\mid \archmath\in\searchspacemath\}$ ideally has high correlation or rank correlation with $\{f(\archmath)\mid \archmath\in\searchspacemath\}$
.
%
In the rest of this section, we give an overview of different types of performance predictors, including learning curve extrapolation (Section \ref{subsubsec:lce}), zero-cost proxies (Section \ref{subsubsec:zcp}), and other methods (Section \ref{subsubsec:subset}).
Note that surrogate models (Section \ref{subsec:bayesian_optimization}) and one-shot models (Section \ref{sec:oneshot}) can also be seen as types of performance predictors.

\begin{figure}
    \centering
    \includegraphics[width=1.0\textwidth]{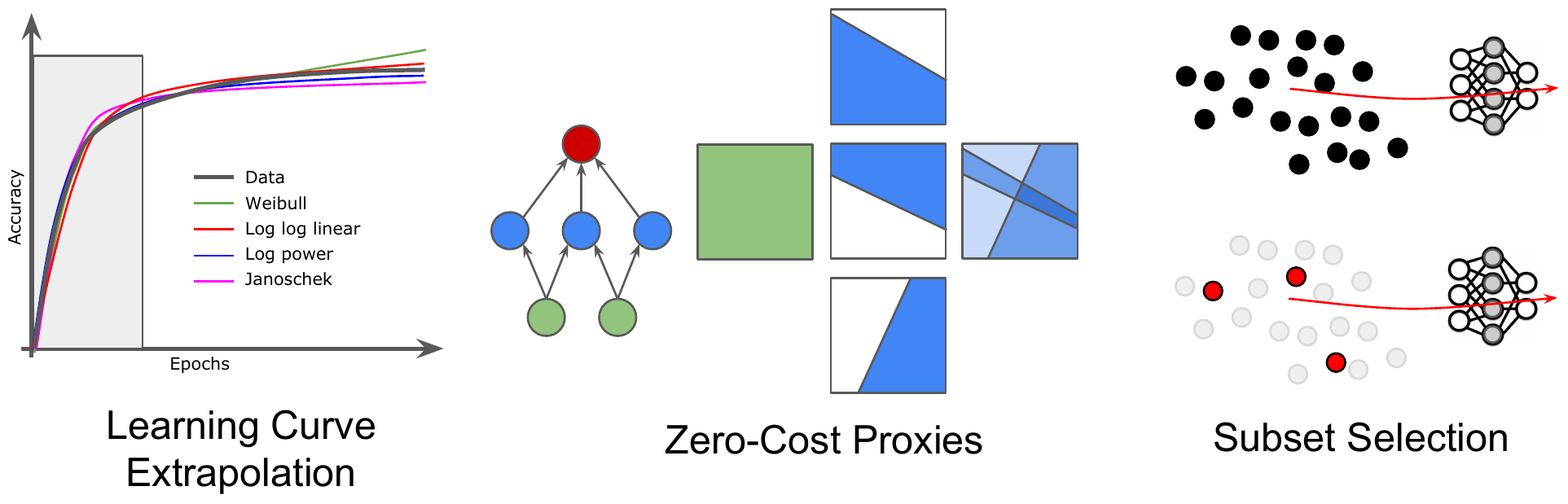} 
    \vspace{-4mm}
    \caption{Illustration of the main types of performance predictors: extrapolating the validation accuracy learning curve via a parameteric model (left), assessing the generalizability of an architecture with a single forward pass of a single minibatch of data (middle), and training the architeture on a subset of the data (right).}
    \label{fig:predictors}
\end{figure}

\subsubsection{Learning Curve Extrapolation} \label{subsubsec:lce}
Learning curve extrapolation methods seek to predict the final performance of a given architecture after partially training it, by extrapolating from its so-called partial \emph{learning curve} (the series of  validation accuracies at all epochs so far).
This can, e.g., be accomplished by fitting the partial learning curve to a parametric model \citep{domhan2015speeding} (see Figure \ref{fig:predictors} (left)).
Learning curve extrapolation methods can also be used together with a surrogate model: in that case, the model takes as input both an encoding of $a$ and a partial learning curve of $a$, and outputs a prediction $f'(a)$ \citep{lcnet,baker2017accelerating}.
Learning curve extrapolation methods can be used to speed up black-box NAS algorithms \citep{domhan2015speeding,ru2020revisiting,nasbenchx11} or in conjunction with multi-fidelity algorithms such as Hyperband or BOHB (described in Section \ref{subsec:multi-fidelity}).

\subsubsection{Zero-Cost Proxies} \label{subsubsec:zcp}
Zero-cost proxies are a recently developed family of performance prediction techniques. 
The idea is to run a very fast computation (such as a single forward and backward pass of a single minibatch of data) over a set of architectures that assigns a score to each architecture, with the hope that the scores are correlated with the final accuracies \citep{mellor2021neural}.
These techniques get their ``zero-cost'' name since the overall time to score each architecture is negligible (often less than 5 seconds) compared to most other performance prediction techniques \citep{abdelfattah2021zerocost}.
While most zero-cost proxies compute architecture scores from a (single) minibatch of data, some are \emph{data-independent}, computing the score solely from the initialized weights or number of parameters of the neural network.

Zero-cost proxies were first introduced by \citet{mellor2021neural}, who estimated the relative performance of neural networks based on how well different linear regions of the network map are separated (see Figure \ref{fig:predictors} (middle)).
Since the initial technique, several new zero-cost proxies have been introduced.
\citet{abdelfattah2021zerocost} made a connection to the pruning-at-initialization literature \citep{lee2019snip, wang2020picking, tanaka2020pruning, theis2018faster} and used this connection to introduce five zero-cost proxies. 
Their best-performing method, \texttt{synflow} \citep{tanaka2020pruning}, is a data-independent method which 
computes the L1 path-norm of the network: it computes the sum of the product of all initialized weights in each path connecting the input to the output.

Since then, two other data-independent methods have been introduced, based on a series of synthetic proxy tasks to test scale invariances and spatial information \citep{li2021generic}, and based on approximating the neural network as a piecewise linear function \citep{lin2021zen}.
Other data-dependent methods make use of the neural tangent kernel (NTK) \citep{jacot2018neural}, based on approximating its trace norm \citep{shu2021nasi} or approximating its spectrum \citep{chen2021neural}.

Although zero-cost proxies have received significant attention since they were first introduced, recent work has shown that simple baselines such as ``number of parameters'' and ``FLOPs'' are surprisingly competitive with all leading techniques.
The main downsides of using zero-cost proxies are that they may be unreliable, especially on larger search spaces \citep{ning2021evaluating, chen2022nasbenchzero,white2022deeper}. 
They also may have biases, such as preferring larger models \citep{ning2021evaluating}
or wide channels \citep{chen2022nasbenchzero}, although the biases can be removed \citep{krishnakumar2022nas}.

On the other hand, recent work encourages the viewpoint that zero-cost proxies are ``weak learners'' which can be combined with other techniques, including other zero-cost proxies, to improve performance \citep{white2022deeper,krishnakumar2022nas}.
Initial work shows that zero-cost proxies can be successfully added to both 
Bayesian optimization-based NAS \citep{white2021powerful, shen2021proxybo} and one-shot NAS \citep{xiang2021zero}.

\subsubsection{Other Low-Fidelity Predictions}\label{subsubsec:subset}

Beside training for fewer epochs, other works give a low-fidelity estimate of the final accuracy by training on a subset of the training data (or a smaller, synthetically generated dataset). This is visualized in Figure \ref{fig:predictors} (right).

Multiple works have studied different subset selection algorithms, such as random sampling, entropy-based sampling \citep{na2021accelerating}, clustering via core-sets \citep{shim2021core}, facility location \citep{prasad2022speeding}, and $k$-center \citep{na2021accelerating}.
\citet{prasad2022speeding} introduce adaptive subset selection to NAS, in which the subset is updated throughout training in order to maximize validation accuracy.

\citet{such2020generative} introduce \emph{generative teaching networks}
which use a small set of synthetic data to train neural networks much
faster than using the original real training data.
The synthetic data is created using a data-generating network to match the accuracy of a network trained on real data.
A related method is \emph{synthetic petri dish} \citep{rawal2020synthetic}, which evaluates architecture motifs by placing them into a small neural network
and then training them using a small synthetic dataset. This latter method also explicitly optimizes the correlation between architecture rankings with the approximation and the full training.



\subsection{Multi-Fidelity Algorithms} \label{subsec:multi-fidelity}
While the previous section was devoted to \emph{methods} of predicting the performance of neural networks, now we cover algorithms that use these methods to run NAS efficiently.


Formally, the objective function $f : \mathcal{X} \longrightarrow \mathcal{R}$, which is typically expensive to fully evaluate, can be cheaply approximated by a lower-fidelity version $\hat{f}(\cdot, b)$ of $f(\cdot)$, parameterized by the fidelity parameter $b$. When $b = b_{max}$, we retrieve the true function $f(\cdot) = \hat{f}(\cdot, b_{max})$. 
This is a generalization of the definition from Section \ref{subsec:prediction}.
The fidelity parameter can denote the number of training epochs, training data subset size, and it can make use of performance prediction techniques from the previous section.
One can even use multiple fidelity parameters at a time \citep{kandasamy17a, zhou2020econas}.
Next, we describe the optimization algorithms that exploit access to multi-fidelity function estimates $\hat{f}(\cdot, b).$

\emph{SuccessiveHalving} (SH) \citep{jamieson16} is one of the simplest multi-fidelity algorithms. 
It starts to train a large number of architectures, slowly killing off more and more architectures which are not promising based on lower fidelity evaluations, until only the most promising architectures are evaluated at the highest fidelity.
The fidelity thresholds and number of architectures to promote to higher fidelities are controlled by a hyperparameter.
A popular improvement to SH is Hyperband (HB) \citep{hyperband}, a multi-armed bandit strategy that repeatedly calls SH as a subroutine, using different values of the minimum budget for each call. 
Therefore, HB hedges its bets against any single choice of the minimum budget.

While SH and HB are purely based on (smart) random search, recent works have combined HB with both Bayesian optimization and evolution.
Bayesian optimization hyperband (BOHB) \citep{bohb,smacv3} works similarly to HB in its first iteration, and on later iterations it fits a probabilistic surrogate model for each fidelity in order to make informed sampling decisions.
Similarly, DEHB \citep{dehb} combines differential evolution \citep{storn_de} with HB, significantly improving the later iterations of HB. 
ASHA \citep{li2018system} and ABOHB \citep{abohb} improve SH and BOHB further, respectively, by making use of massively parallel asynchronous computation and early stopping strategies. 
Finally, EcoNAS \citep{zhou2020econas} proposes a hierarchical evolutionary search method that partitions the search space into subsets and allocates increasing fidelities to the most promising architectures in each subset.

\subsection{Meta-Learning} \label{subsec:meta-learning}

A majority of NAS approaches consider solving a single task from scratch, ignoring previously explored solutions. However, this is in contrast to what both researchers and practitioners typically do. 
Often, architectures are transferred across datasets and even across tasks, and on a new task, researchers typically start with a state-of-the-art solution. 
So, one might ask: why run NAS from scratch rather than re-using information from, e.g., previous experiments? 
This question naturally leads to the idea of \emph{meta-learning} or \emph{learning to learn} \citep{schmidhuber:1987:srl, Thrun1996LearningTL,HochreiterL2L}, which aims at improving a learning algorithm by leveraging information from past, related experiments \citep{vanschoren_meta_2019,Hospedales_ml_survey}. 


\citet{wong_neuralautoml} and \citet{zimmer-tpami21a} employ meta-learning strategies in a more general automated machine learning setting. 
Since the focus is not on NAS, they both solely consider a small set of candidate architectures. 
In \citet{wong_neuralautoml}, tasks are encoded in a similar fashion as word embeddings in NLP \citep{NIPS2013_9aa42b31}. 
In contrast, \citet{zimmer-tpami21a} simply warm-start their search based on previously well-preforming configurations.

\citet{anonymous2020towards} and \citet{Elsken20} focus on few-shot learning: the problem of learning a new task with just a few data points for training. 
The authors extend gradient-based, model-agnostic meta-learning approaches such as MAML \citep{finn_maml} and REPTILE \citep{reptile} to not only meta-learning an initial set of weights for a fixed neural network architecture, but also to the architecture itself by incorporating a differentiable method such as DARTS \citep{darts} into the meta-learning algorithm.

The work by \citet{lee2021rapid} is neither restricted to few-shot learning nor to choosing architectures from a small set of candidates. 
Rather, they employ typical NAS search spaces such as the ones discussed in Section \ref{sec:search_spaces}. 
The authors propose a novel set encoder to improve upon deep sets \citep{zaheer_deepsets} and set transformers \citep{lee2019set}. 
A graph neural network-based decoder is employed to generate neural architectures given a set encoding. 
Additionally, a graph neural network is employed to encode generated architectures. The architecture encoding in combination with the set encoding is then used to meta-learn a surrogate model to predict the performance of the architecture, dataset tuple. \citet{shala2022transfer} extend the work by \citet{lee2021rapid} by employing the dataset and architecture encodings within a Bayesian optimization framework, resulting in a probabilistic surrogate predictor. This further enables adapting the surrogate to datapoints seen at test time.

\subsection{Weight Inheritance and Network Morphisms} \label{subsec:inheritance}

While black-box optimization-based NAS algorithms train each architecture from scratch, and one-shot methods train \emph{all} architectures with the same set of weights, a line of work proposes an in-between solution: reuse the weights of trained architectures on similar untrained architectures.
This idea is especially helpful for black-box optimization approaches that apply only small, sequential changes to architectures when generating a new candidate architecture. 
For example, \citet{Real17} propose to copy the weights of all layers that have not been affected by applied mutations from the parent architecture to its offspring. 

This idea has also been extended by the concept of \emph{network morphisms} \citep{Chen16net2net,wei_netmorph}. 
Network morphisms are operators acting on the space of neural network architectures. 
They change the architecture of a neural network without changing the function they represent, i.e., given an arbitrary input, the output remains identical for the original architecture and the architecture having been modified by a network morphism. 
This is typically achieved by properly initializing the modified architecture. 
Network morphisms have been employed in evolutionary algorithms \citep{elsken2017simple, Elsken19, wistuba_netmorph,schorn2019automated}, reinforcement learning \citep{cai2018efficient, cai_path-level_2018}, Bayesian optimization \citep{auto-keras}, and even one-shot methods \citep{Fang2020Fast}.


\section{Extensions} \label{sec:extensions}

The previous sections studied the main techniques from the classic instantiation of NAS.
In this section, we survey a few common extensions: joint NAS + HPO, constrained/multi-objective NAS, and neural ensemble search.

\subsection{Joint NAS + HPO} \label{sec:nas_hpo}

While a large body of the NAS literature assumes fixed hyperparameters in their experimental setup, it has been shown -- perhaps not very surprisingly -- that hyperparameters also play a significant role. 
For example, on the DARTS search space, tuning hyperparameters can lead to a huge improvement, exceeding the performance gains obtained by NAS \citep{yang2019evaluation}.
However, the best hyperparameters may vary significantly across architectures even in the same search space \citep{yang2019evaluation}.
Therefore, a recent body of work seeks to overcome these challenges and give efficient algorithms for NAS + HPO \citep{zela2018towards, dong2020autohas, dai2021fbnetv3,izquierdo2021bag,zhou2021dha}.

Running joint NAS + HPO is significantly more challenging than running NAS or HPO in isolation. First, the complexity of the search space is substantially increased, due to the increased number of hyperparameters and the heterogeneity of the hyperparameters.
Second, the interaction between architectures and training hyperparameters in terms of network performance is difficult to model. 
Furthermore, some hyperparameters can have different effects on the performance under different evaluation budgets, reducing the effectiveness of many multi-fidelity and performance prediction techniques.

In light of these challenges, several solutions have been proposed. 
Various methods have been introduced to homogenize the search space, such as reformulating NAS as an HPO problem with categorical hyperparameters \citep{zela2018towards}, or standardizing the representation of the NAS and HPO hyperparameters by assigning continuous-valued coefficients in $[0,1]$ \citep{dong2020autohas}.
The search strategies resemble standard NAS algorithms such as BO \citep{zela2018towards,dai2021fbnetv3,izquierdo2021bag}, evolution \citep{dai2021fbnetv3,izquierdo2021bag}, or REINFORCE with weight sharing \citep{dong2020autohas}.

\subsection{Constrained and Multi-Objective NAS} \label{subsec:hardware}

Although NAS has been very popular in recent years, most work focuses on solely optimizing for a single objective, typically 
the accuracy or error rate. However, there are many settings for which this is not sufficient, such as when the neural network must be deployed on an edge device or must satisfy a legal definition of fairness.
In such applications, we may need to constrain the latency, memory usage, or rate of errors across classes \citep{sukthanker2022importance}.
There has been particular interest in constraints related to edge devices and other hardware, termed \emph{hardware-aware NAS} \citep{benmeziane2021comprehensive}.
To achieve one or more objectives in addition to accuracy, the standard NAS objective is typically modified to either a \emph{constrained} optimization problem (e.g., \citet{mnasnet, proxylessnas,Bender_2020_CVPR}) or a \emph{multi-objective} optimization problem (e.g., \citet{Elsken19,hu2019efficient,lu_nsganet,lu_nsganet_v2,izquierdo2021bag}).

In constrained optimization, one tries to solve the following equation:
\begin{equation} \label{eq:constrained}
 \min_{\archmath \in \searchspacemath} \nasobjectivemath(\archmath) \text{  subject to  } h_i(\archmath) \le c_i \text{ for } \: i \in \{1, \dots, k\}
\end{equation}
where \nasobjective denotes, as before, the original objective function (e.g., validation error), and $h_i$ represent hardware constraints as a function of the architecture. 
This problem is often solved by a transform into an additive or multiplicative unconstrained problem such as 
$\min_{\archmath \in \searchspacemath} \nasobjectivemath(\archmath) + \sum_i \lambda_i g_i(\archmath)$ 
with  penalty functions $g_i$ penalizing architectures not satisfying the constraints, e.g., $g_i(\archmath) = \max \big(0,h_i(\archmath) - c_i\big)$ and hyperparamters $\lambda_i$ trading off the objectives and constraints. 
This single-objective optimization problem is then solved using black-box optimization methods or one-shot methods.
In the latter case, the penalty functions $g_i$ needs to be differentiable, which is often not the case.
Therefore, discrete metrics such as latency are relaxed to continuous variables through various techniques, such as with a Gumbel softmax function \citep{fbnet}.

In multi-objective optimization, the requirements in Equation \ref{eq:constrained} are treated as separate objectives that are optimized along with the original objective:
\begin{align*}
& \min_{\archmath \in \searchspacemath}  \Big( \nasobjectivemath(\archmath), h_1(\archmath), \dots, h_k(\archmath)  \Big).
\end{align*}
While this can again be reduced to a single-objective problem via scalarization methods, another common approach is to search for a set of \emph{non-dominated} solutions that are optimal in the sense that one cannot reduce any objective without increasing at least one other objective. 
The set of non-dominated solutions is called the \emph{Pareto front}.
The most common approach in this case is to employ multi-objective evolutionary algorithms which maintain a population of architectures and aim to improve the Pareto front obtained from the current population by evolving the current population \citep{Elsken19,hu2019efficient,lu_nsganet,izquierdo2021bag}. 
Multi-objective evolutionary algorithms have also been used in combination with weight sharing within one-shot models \citep{lu_nsganet_v2, supernetwork_generation}.

One of the most widely-studied constrained NAS problems is regarding hardware efficiency such as memory or latency, and many works have been devoted to efficiently approximating hardware metrics of interest.
While simple metrics such as number of parameters are easily computed, these are often not correlated enough with other metrics of interest such as memory or latency.
Other solutions include computing hardware costs modularly as the sum of the hardware cost of each operation \citep{proxylessnas} or by using a surrogate model that predicts hardware costs \citep{brpnas,laube2022what}.

\subsection{Neural Ensemble Search} \label{sec:ensemble}

While the goal of neural architecture search is to return the best standalone architecture, ensembling methods are popular within the deep learning community for their robust predictions and their easy uncertainty quantification.
A newly emerging extension of NAS is concerned with finding the best \emph{ensemble} of neural networks with diverse architectures, which can outperform standard NAS in terms of accuracy, uncertainty calibration, and robustness to dataset shift \citep{zaidi2021neural}.
Neural ensemble search is defined as follows:
\begin{align}
    \min_{a_1, \dots, a_M \in \archss}  & \lval \left(\ens \left( (w^*(a_1),a_1), \dots, (w^*(a_M),a_M)\right) \right) \label{eq:nes-optim} \\
    &\text{s.t.}\quad w^*(a) = \text{argmin}_w~ \ltrain\left(w,a\right)~~ \forall a\in\archss, \nonumber
\end{align}
where $\ens$ is the function which aggregates the outputs of $f_1, \dots, f_M$.
Note that the search space cardinality is $|\archss|^M$ rather than $|\archss|$ as in standard NAS.

\citet{zaidi2021neural} propose two simple yet effective procedures based on random search and regularized evolution \citep{real2019regularized} that search for architectures that optimize Equation \ref{eq:nes-optim}. 
Despite their effectiveness, these algorithms take considerable computation due to the black-box nature of the optimization algorithms. 
Multi-headed NES \citep{Narayanan2021b} circumvents this issue by applying differentiable NAS methods on the heads of a multi-headed network. 
The heads are explicitly tuned to optimize the ensemble loss together with a diversity component that encourages uncorrelated predictions coming from the individual heads. 
Other works have set up neural ensemble search with a one-shot model for the entire architecture.
NESBS (Neural Ensemble Search via Bayesian Sampling) \citep{Shu2021NeuralES} propose to use a supernet to estimate the ensemble performance of independently trained base learners and then use Bayesian sampling to find a high-performing ensemble.
NADS (Neural Architecture Distribution Search) \citep{Ardywibowo2020NADSNA} follows a similar line by training a supernet to optimize an objective that is tailored to provide better uncertainty estimates and out-of-distribution detection. 
\citet{Chen2021OneShotNE} run evolutionary search on the supernet to find a high-performing ensemble.

\section{Applications} \label{sec:applications}

Along with discovering improved architectures for well-known datasets,
one of the primary goals of the field of NAS is to quickly and automatically
find high-performing architectures for brand new datasets and tasks.
Although the majority of the NAS literature focuses on image classification,
there are numerous success stories for NAS applied to less well-known settings.
In this section, we discuss a few of these successes, including graph neural networks, generative adversarial networks, dense prediction, and transformers.

\subsection{Graph Neural Networks}

Graph neural networks (GNNs) are designed to process data represented by graphs. 
Using NAS to design GNNs poses unique problems: the search space for GNNs is more complex than typical convolutional search spaces, and
both NAS and GNNs are independently known for their large computational overhead.

\citet{zhou2019auto} initiated a line of work applying NAS to GNNs by defining a new search space with GNN-specific operations and then using a reinforcement learning strategy.
Follow-up work designed similar search spaces \citep{gao2020graph, zhang2021automated}.
with specialized features such as meta-paths \citep{ding2021diffmg}, 
edge features \citep{jiang2020graph}, or fast sampling operations \citep{gao2020graph}.

Overall, the main difference between NAS for GNNs and more standard NAS settings lies in the construction of the search space. 
The main search strategies used by GNN NAS algorithms are typical NAS approaches:
reinforcement learning \citep{zhou2019auto, gao2020graph, zhao2020simplifying}, one-shot methods \citep{ding2021diffmg, zhao2020probabilistic}, and evolutionary algorithms \citep{jiang2020graph, nunes2020neural}.
For a detailed survey on NAS for GNNs, see \citet{zhang2021automated}.



\subsection{Generative Adversarial Network}
Generative adversarial networks (GANs) \citep{goodfellow2014generative} are a popular choice for generative modeling in tasks such as computer vision. 
GANs make use of two separate networks training in tandem: a generator and a discriminator. 
Due to having two separate networks, and their notoriously brittle training dynamics \citep{gulrajani2017improved}, 
GANs require special techniques for effective NAS.

Different works have achieved improved performance via NAS by searching for only the generator architecture with a fixed discriminator \citep{doveh2021degas}, with a predefined progressively growing discriminator \citep{fu2020autogan}, or by searching both the generator and discriminator architectures simultaneously \citep{gong2019autogan}.
The most popular choice of search space is the cell-based search space.
The cell for the generator consists of a standard convolutional cell, with the addition of various upsampling operations \citep{ganepola2021automating, gong2019autogan, tian2020off}.

The search techniques resemble the techniques used for standard NAS: reinforcement learning \citep{fu2020autogan, wang2019agan, tian2020off}, one-shot NAS \citep{gao2020adversarialnas, doveh2021degas, lutz2018alphagan}, and evolutionary algorithms \citep{kobayashi2020multi}, with scoring based on either Inception Score (IS) \citep{salimans2016improved} or
Fr\'{e}chet Inception Distance (FID) \citep{heusel2017gans}.
For a comprehensive survey on NAS for GANs, see \citet{ganepola2021automating}.

\subsection{Dense Prediction Tasks}
Dense prediction for computer vision encompasses a variety of popular tasks such as semantic segmentation, object detection, optical flow, and disparity estimation, and it requires more complex architectures compared to standard image classification problems.
For example, the architectures often include a decoder \citep{unet}, modules for generating multi-scale features \citep{aspp} or task-specific heads \citep{rcnn} in addition to the main network. 
Thus, NAS algorithms have been applied to search for these components, either in isolation \citep{dpc,Ghiasi_2019_CVPR,Xu_2019_ICCV} or jointly \citep{Guo_2020_CVPR,yao_2020_aiii}, or by discovering novel design patterns \citep{Du_2020_CVPR}. 
For a  survey on NAS for dense prediction, see \citet{elsken2022neural}.

Once again, standard NAS techniques are used:
\citet{Liu_2019_CVPR,Xu_2019_ICCV,Saikia_2019_ICCV,Guo_2020_CVPR} employ gradient-based search via DARTS \citep{darts}; \citet{Ghiasi_2019_CVPR, Du_2020_CVPR} use RL; \citet{Bender_2020_CVPR} is inspired by ProxylessNAS \citep{proxylessnas} and ENAS \citep{pham2018efficient}.

Methods for dense prediction tasks (e.g., \citet{Shaw_2019_ICCV,Wu_2019_CVPR,Bender_2020_CVPR,NIPS2019_8890,Guo_2020_CVPR}) typically build search spaces based on state-of-the-art image classification networks, with task-specific components from well-performing dense prediction architecture components.
As many approaches fix the backbone and only search for other task-specific components of the architecture, they often employ pre-trained backbone architectures \citep{Guo_2020_CVPR,Chen_2020_CVPR} or even cache the features generated by a backbone \citep{dpc, Wang_2020_CVPR,Nekrasov_2019_CVPR} to speed up architecture search. 
\citet{dpc, Ghiasi_2019_CVPR} also use a down-scaled or different backbone architecture during the search process. 
Methods also sometimes employ multiple search stages, with the goal of first eliminating poorly performing architectures (or parts of the search space) and successively improving the remaining architectures \citep{Guo_2020_CVPR,Du_2020_CVPR}.

Overall, while it is much harder to run NAS on dense prediction tasks compared to image classification tasks because of the computational demands of dense prediction, there has been a rapid increase in developments with the rise of computationally efficient one-shot NAS methods.
While efforts thus far have focused on semantic segmentation and object detection, avenues for future work include disparity estimation, panoptic segmentation, 3D detection and segmentation, and optical flow estimation.


\subsection{Transformers} \label{subsec:transformers}

Transformers were proposed by \citet{vaswani2017attention} to help with the issue of longer sequences that RNNs had difficulty modeling, by using self-attention and cross-attention mechanisms such that each token's representation in an input sequence is computed from a weighted average of the representation of all other tokens. 
The core transformer design was introduced for machine translation, but it has found widespread usage in causal language modeling \citep{gpt2,gpt3}, masked language modeling \citep{bert,clark2020electra,liu2019roberta}, and more recently, computer vision \citep{dosovitskiy2020image, liu2021swin}.
Since its release, there have been many efforts to improve transformers via NAS. 
The most common search strategies for transformers are evolutionary \citep{so2019evolved,chen2021autoformer,primer} or one-shot \citep{gong2021nasvit,su2021vitas,li2021bossnas,ding2021hr} 
On the other hand, there is a huge variety of different search spaces that have been tried recently, relative to other areas (e.g., in NAS for convolutional architectures, the majority of works use cell-based search spaces). 
Overall, the field of NAS for transformers has not converged to one ``best'' type of search space.
Below, we survey NAS methods for four types of transformers: decoder-only, encoder-only, encoder-decoder, and vision transformers.
See \citet{chitty2022neural} for an in-depth survey.

Decoder-only architectures, such as the GPT line of architectures \citep{gpt2,gpt3} directly consume the input text prompt and output the sequence of text tokens that are most likely 
to follow.
Primer \citep{primer} is a NAS algorithm that makes use of evolutionary search on a large macro decoder-only search space.
The approach found two consistent improvements to the transformer block: squaring the \texttt{ReLU} in the feedforward block in the transformer layer, and adding depthwise convolutions after self-attention heads.

Encoder-only architectures, such as BERT \citep{bert} encode the input text into a representation which can be used for many kinds of downstream tasks.
Multiple works \citep{xu2022analyzing,nasbert,autotinybert} seek to discover compressed versions of BERT, in which the desired latency and task are specified by the user. 
The typical approach is to train a supernet on a standard self-supervised task (masked language modeling), which can then be used to discover compressed models for a given language task.

Encoder-decoder architectures such as T5 \citep{raffel2020exploring} are used in sequence-to-sequence tasks such as machine translation, in which the source language is encoded into a representation, which is then decoded into the target language.
\citet{so2019evolved} use evolutionary search together with a new technique to dynamically allocate more resources to more promising candidate models, while
\citet{zhao2021memory} propose a DARTS-based algorithm with a new technique for memory efficiency in backpropagation.
Finally, KNAS \citep{xu2021knas} and SemiNAS \citep{seminas} speed up the search using zero-cost proxies and a surrogate transformer model, respectively.

A large variety of NAS algorithms have been studied for vision transformer search spaces, with the majority using one-shot methods.
AutoFormer \citep{chen2021autoformer} searches over vision transformer architectures and hyperparameters using a single-path-one-shot strategy \citep{guo2020single} and then running evolutionary search on the trained supernet.
A followup work, AutoFormerv2 \citep{chen2021searchingthesearchspace}, automated the design of the search space itself by gradually evolving different search dimensions.
Other works have improved supernet training via gradient conflict aware training \citep{gong2021nasvit} or channel-aware training \citep{su2021vitas}.
Finally, \citet{li2021bossnas} and \citet{ding2021hr} run one-shot methods on hybrid CNN and transformer search spaces for computer vision.


\section{Benchmarks} \label{sec:benchmarks}

In the early days of NAS research, the most popular metrics were the final test accuracies on CIFAR-10 and ImageNet. 
This caused inconsistent search spaces and training pipelines across papers, and also drove up computational costs. 
For example, it became standard to train the final architecture for 600 epochs, even though the test accuracy only increases by a fraction of a percent past 200 epochs.
Recently, queryable NAS benchmarks have helped the field reduce computation when developing NAS techniques and to achieve fair, statistically significant comparisons between methods.

A \emph{NAS benchmark} \citep{lindauer2019best} is defined as a dataset with a fixed train-test split, a search space, and a fixed evaluation pipeline for training the architectures.
A \emph{tabular} NAS benchmark is one that additionally gives \emph{precomputed evaluations} for all possible architectures in the search space.
A \emph{surrogate} NAS benchmark is a NAS benchmark along with a surrogate model that can be used to predict the performance of any architecture in the search space.
A NAS benchmark is \emph{queryable} if it is either a tabular or a surrogate benchmark.
Queryable NAS benchmarks can be used to efficiently simulate many NAS experiments using only a CPU, by querying the performance of neural networks from the benchmark, rather than training them from scratch.
In the rest of the section, we give an overview of popular NAS benchmarks. 
See Appendix Table \ref{tab:benchmarks_comprehensive} for a summary.

The first tabular NAS benchmark was NAS-Bench-101 \citep{nasbench}. It consists of a cell-based search space of $423\,624$ architectures, each with precomputed validation and test accuracies on CIFAR-10 for three different seeds.
A follow-up work, NAS-Bench-1Shot1 \citep{nasbench1shot1}, is able to simulate one-shot algorithms by defining subsets of the NAS-Bench-101 search space which have a fixed number of nodes.
NAS-Bench-201 \citep{nasbench201} is another popular tabular NAS benchmark, consisting of $6466$ unique architectures, each with precomputed validation and test accuracies on CIFAR-10, CIFAR-100, and ImageNet-16-120 for three seeds each.
NATS-Bench \citep{natsbench} is an extension of NAS-Bench-201 which also includes a macro search space.
Another extension, HW-NAS-Bench-201 \citep{li2021hwnasbench}, gives the measured or estimated hardware cost for all architectures across six hardware devices.

Surr-NAS-Bench-DARTS (formerly called NAS-Bench-301) \citep{nasbench301} was the first surrogate NAS benchmark, created by training $60\,000$ architecture from the DARTS \citep{darts} search space on CIFAR-10 and then training a surrogate model.
The authors also released Surr-NAS-Bench-FBNet for the FBNet search space \citep{fbnet}.
A follow-up work, NAS-Bench-x11 \citep{nasbenchx11}, devised a technique to predict the full learning curve, allowing the validation accuracies to be queried at arbitrary epochs, which is necessary for simulating multi-fidelity NAS algorithms.
%

TransNAS-Bench-101 \citep{transnasbench} is a tabular benchmark that covers seven different computer vision tasks from the Taskonomy dataset \citep{zamir2018taskonomy}.
Beyond computer vision, NAS-Bench-NLP \citep{nasbenchnlp} consists of an LSTM-inspired search space for NLP, and NAS-Bench-ASR \citep{nasbenchasr} is a tabular NAS benchmark for automatic
speech recognition \citep{garofolo1993timit}.
NAS-Bench-360 \citep{nasbench360} is a benchmark suite which gives NAS benchmarks on ten diverse problems such as prosthetics control, PDE solving, protein folding, and astronomy imaging, and is search space agnostic, although three of the tasks have pretrained architectures on the NAS-Bench-201 search space.
Finally, NAS-Bench-Suite \citep{nasbenchsuite} is a benchmark suite which combines the majority of existing queryable NAS benchmarks, 28 total tasks, into a single unified interface. An extension, NAS-Bench-Suite-Zero, offers precomputed zero-cost proxy values across all tasks \citep{krishnakumar2022nas}.

Using queryable benchmarks allows researchers to easily simulate hundreds of trials of the algorithms with different initial random seeds, making it easy to report statistically significant comparisons.
However, over-reliance on a few benchmarks can lead to the field over-fitting \citep{raji2021ai,koch2021reduced} and is not conducive to the discovery of truly novel methods.
Therefore, researchers should use a large set of diverse NAS benchmarks whenever possible.

\section{Best Practices} \label{sec:best_practices}

The field of NAS has at times seen problems with reproducibility and fair, statistically significant comparisons among methods. 
These issues impede the overall research progress in the field of NAS.
Recently, a few papers have laid out best practices and guidelines for conducting sound NAS research
that is reproducible and makes fair comparisons \citep{randomnas, yang2019evaluation, lindauer2019best}.
These best practices are also available as a checklist \citep{lindauer2019best}.
%
We encourage NAS researchers to follow the checklist and to attach it to the appendix of their papers.
Now, we summarize these best practices for NAS research.

\subsection{Releasing Code and Important Details}
It is nearly impossible to reproduce NAS methods without the full code. Even then, random seeds should be specified and reported.
Furthermore, releasing easy-to-use code can lead to more follow-up methods and impact.
For example, \citet{darts} released easy-to-use code for DARTS, which facilitated numerous follow-up works.

When releasing code, it is important to release all components, including the training pipeline(s), search space, hyperparameters, random seeds, and the NAS method.
Many papers use different architecture training pipelines during the search and during the final evaluation, so it is important to include both.
Note that using popular NAS benchmarks such as NAS-Bench-101 or NAS-Bench-201 (see Section \ref{sec:benchmarks}) makes this substantially easier: the training pipeline is already fixed.

NAS methods often have several moving parts.
As a result, they typically have many hyperparameters of their own that could be tuned. 
In fact, many NAS methods themselves make use of neural networks -- one could even run a NAS algorithm on the NAS algorithm!
Due to this complexity, it is important to report if, or how, these hyperparameters were tuned.
When reporting results on a large set of search spaces and datasets, the best practice is to tune the 
hyperparameters of the NAS method on one dataset, and then fix these hyperparameters for the remaining evaluations on other datasets.
We also note that, in general, devising NAS methods with fewer hyperparameters is more desirable, especially because it has recently been shown that hyperparameters often do not transfer well across datasets and search spaces \citep{nasbenchsuite}.

\subsection{Comparing NAS Methods}


When comparing NAS methods, it is not enough to use the same datasets. 
The exact same NAS benchmarks must be used: a dataset with a fixed train-test split, search space, and evaluation pipeline.
Otherwise, it is unclear whether a difference in performance is due to the NAS algorithm or the training pipeline.

Several papers have shown that simple baselines are competitive with state-of-the-art NAS algorithms \citep{sciuto2019evaluating, randomnas, white2021exploring, ottelander2020local}.
When desigining a new method for NAS, it is important to compare the method with baselines such as random sampling and random search.
Furthermore, many NAS methods are anytime algorithms: a time budget does not necessarily need to be specified upfront, and the method can be stopped at any time, returning the best architecture found so far.  
The longer the NAS method runs, the better the final result.
These NAS methods should be compared on a plot of \emph{performance over time}.
Even one-shot algorithms can be compared in this way, since the supernet can be discretized and trained at any point.

We recommend that NAS researchers run thorough ablation studies to show which part(s) of the NAS method lead to the most improved performance. 
As mentioned in the previous section, NAS methods often have several moving parts, so a clean understanding of the importance of each part and how they work together, is important to report.
Finally, we recommend that researchers run multiple trials of their experiments and report the random seeds for each experiment. 
NAS methods can have high variance in the randomness of the algorithm, so running many trials is important to verify statistically significant comparisons.

\section{Resources} \label{sec:resources}

In this section, we discuss NAS resources including libraries (Section \ref{subsec:libraries}), other survey papers (Section \ref{subsec:surveys}), and additional resources (Section \ref{subsec:additional_resources}).

\subsection{Libraries} \label{subsec:libraries}

A long line of engineering has been focused on automating machine learning pipelines: Auto-WEKA \citep{thornton2013autoweka}, Auto-Sklearn \citep{feurer2015autosklearn}, TPOT \citep{olson2016gecco}, and AutoGluon-Tabular \citep{erickson2020autogluon}. 
More recently, a special focus has been given to developing tools that can facilitate the deployment of various NAS algorithms for practitioners, such as Auto-Keras \citep{Jin2019AutoKeras}, Auto-PyTorch Tabular \citep{zimmer-tpami21a}, AutoGluon \citep{erickson2020autogluon}, and NNI \citep{nni2021}.

To provide a toolbox for facilitating NAS research, in both developing new NAS methods and applying NAS to new problem domains, various libraries have been proposed.
The DeepArchitect library \citep{negrinho2017deeparchitect}, which separates the search space from the optimizer, was an important first step towards this direction in the NAS community.
NASLib \citep{ruchte2020naslib} unifies and simplifies NAS research by having a single abstraction for one-shot and BBO algorithms, and a single abstraction for the search spaces of nearly all queryable NAS benchmark.
Archai \citep{hu2019efficient} also provides unified abstractions for one-shot and discrete NAS algorithms. 
The aim for Archai is both to support reproducible rapid prototyping for NAS research as well as to be a turnkey solution for data scientists looking to try NAS on their tasks. 
PyGlove \citep{peng2020pyglove} introduced a novel approach to constructing NAS methods via symbolic programming, in which the ML programs are mutable and can be manipulated and processed by other programs.

\subsection{Other NAS Survey Papers} \label{subsec:surveys}


There are several older NAS survey papers.
\citet{nas-survey} provides a compact introduction to NAS and introduces the ``three pillars'' of NAS: search space, search strategy, and performance evaluation strategy.
The survey by \citet{wistuba2019survey} provides a more comprehensive view of the landscape of NAS research, unifying and categorizing existing methods. 
\citet{ren2020comprehensive} gave a layout that focused on the historical challenges in the field of NAS, as well as the solutions found to remedy these challenges.

Other surveys have been released which focus on a specific sub-area of NAS.
\citet{liu2021survey} focus on evolutionary NAS, \citet{benmeziane2021comprehensive} focus on hardware-aware NAS (HW-NAS), 
\citet{zhang2021automated} survey AutoML (with a NAS focus) on graphs, \citet{elsken2022neural} survey NAS for dense prediction in computer vision, and \citet{xie2021weight}, \citet{santra2021gradient}, and \citet{cha2022supernet} all survey one-shot NAS methods.

Finally, there are more survey papers with a broader focus such as automated machine learning (AutoML) or automated deep learning (AutoDL), which devote a section to NAS \citep{yao2018taking,he2021automl,yu2020hyper,kedziora2020autonoml,dong2021automated}. 
Notably, the first book on automated machine learning (which is open-access) was released in May 2019 by \citet{automl}.

\subsection{Additional Resources} \label{subsec:additional_resources}

There are multiple long-running workshops which focus on NAS and related topics. 
The AutoML workshop at ICML (2014-2021)
and Meta-Learning workshop at NeurIPS (2017-2022)
have had a healthy overlap in attendance with the NAS community, especially over the last few years, while ICLR (2020, 2021) and CVPR (2021) have had workshops devoted solely to NAS.
Finally, after many years of AutoML and NAS workshops, the community has grown large enough to start the first AutoML conference: \url{https://automl.cc/}.

For a continuously updated, searchable list of NAS papers, see \url{https://www.automl.org/automl/literature-on-neural-architecture-search/}. 
For a continuously updated list of NAS papers published at ML venues, as well as other resources, see \url{https://github.com/D-X-Y/Awesome-AutoDL}.

\section{Future Directions} \label{sec:future_directions}

Neural architecture search has come a long way in the last few years.
The efficiency of NAS algorithms has improved by orders of magnitude, tools exist to compare NAS algorithms without GPUs, and researchers have created many novel techniques and diverse search spaces. 
Architectures discovered by NAS constitute the state of the art on many tasks.
However, there are still many unsolved problems and promising future directions. 
In this section, we discuss a few of the most important directions for future work in NAS.

\subsection{Robustness of Efficient Methods}

One-shot methods are one of the most popular techniques for NAS due to their orders-of-magnitude speedups over to black-box optimization techniques.
While one-shot techniques have already seen major progress, they still face performance issues.

Even though many improvements of one-shot algorithms such as DARTS have been proposed (see Section \ref{subsec:darts}), these works generally focus on a single improvement; the field lacks a large-scale, fair comparison among one-shot methods.
Furthermore, as it currently stands, applying one-shot methods to a new task requires a significant amount of expertise.
Devising one-shot approaches that work robustly and reliably across new datasets and tasks is an important area for future study.

Another more recent set of techniques that promises orders-of-magnitude speedups are zero-cost proxies (see Section \ref{subsubsec:zcp}).
Although recent work has shown that many zero-cost proxies do not consistently outperform simple baselines \citep{ning2021evaluating}, other work argues that there is untapped potential for zero-cost proxies \citep{white2022deeper}, especially when combined with existing NAS techniques \citep{white2021powerful, xiang2021zero}.
Developing a better understanding of \emph{when} and \emph{why} zero-cost proxies work in certain settings is an important area for future research.

\subsection{Going Beyond Hand-Crafted, Rigid Search Spaces}

The search spaces for NAS methods are typically carefully hand-designed by human experts. 
While carefully designing search spaces decreases search times, it also contradicts the idea of having an automated system that can be employed by non-experts, and it limits the scope of NAS to domains where strong search spaces are available. 
Furthermore, in the last few years, the most-studied type of search space by far has been the cell-based search space, which is significantly more rigid than other types of search spaces.

Hierarchical search spaces offer a better trade-off between flexibility and ease of search, yet they are relatively under-explored when compared to cell-based search spaces (see Section \ref{subsubsec:hierarchical}).
Furthermore, hierarchical search spaces by nature have a higher diversity when compared to cell-based search spaces, reducing the overall human bias of the search space.

Optimizing search spaces in an automated manner \citep{ru2020nago} such as starting with large, diverse search spaces and then iteratively pruning low-performing parts of the space \citep{Radosavovic,Guo_2020_CVPR} could allow researchers to consider a significantly larger variety of architectures.

\subsection{Fully Automated Deep Learning}
Although NAS has seen a huge amount of interest, recent work has shown that on popular search spaces such as the DARTS search space, optimizing the training hyperparameters leads to a greater increase in performance than optimizing the architecture \citep{yang2019evaluation, nasbench1shot1}. 
While these results show that for some search spaces, optimizing hyperparameters may be more important than optimizing the architecture, the best case scenario is to optimize both hyperparameters and the architecture simultaneously.

A new thread of research seeks to simultaneously optimize the hyperparameters and architecture: NAS + HPO (see Section \ref{sec:nas_hpo}). 
Varying hyperparameters along with the architecture also significantly reduces human bias, making it possible to discover previously unknown combinations of architectures and hyperparameters that substantially outperform existing methods.
Therefore, while this problem is significantly more challenging than NAS or HPO alone, the potential improvements are much higher. 

Furthermore, we do not need to stop just at NAS + HPO: we can optimize the full deep learning pipeline, including problem formulation, data processing, data augmentation, model deployment, and continuous monitoring. 
In other words, the goal is to run fully automated deep learning (AutoDL) \citep{dong2021automated}.
As the field of NAS matures, AutoDL has the potential to play a big role in realizing substantial improvements in performance for real-world problems.

\acks{
We thank Difan Deng and Marius Lindauer for creating and maintaining their literature list on neural architecture search, which has made this survey significantly easier to write \citep{deng-21}.
This research was partially supported by TAILOR, a project funded by EU Horizon 2020 research and innovation programme under GA No 952215.
We acknowledge funding by European Research Council (ERC) Consolidator Grant ``Deep Learning 2.0'' (grant no.\ 101045765). Funded by the European Union. Views and opinions expressed are however those of the author(s) only and do not necessarily reflect those of the European Union or the ERC. Neither the European Union nor the ERC can be held responsible for them.
\begin{center}
\includegraphics[width=0.3\textwidth]{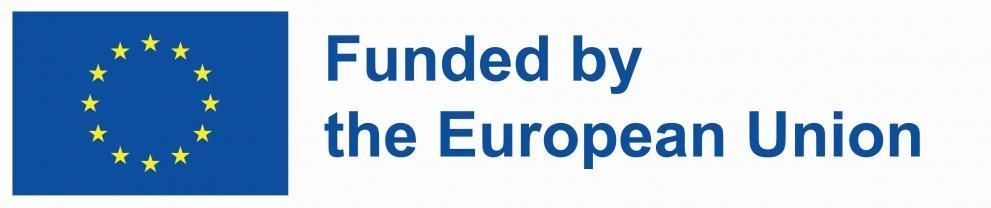}
\end{center}
}

\clearpage

\appendix

\section{Additional Figures and Tables}

\begin{figure}
    \centering
    \includegraphics[trim=2cm 9cm 2cm  1cm, clip, width=1.0\linewidth]{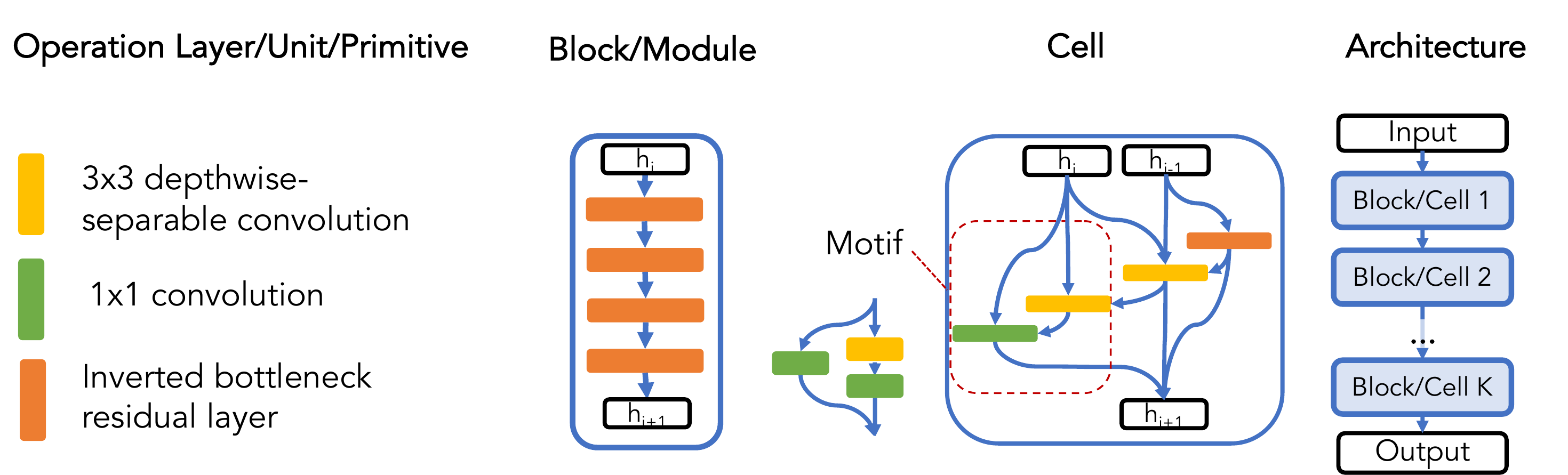}
    \caption{NAS search space terminology.
    Operation layers/units/primitives consist of sets of 1-3 operations.
    A block/module denotes a sequential stack of layers in chain-structured or macro search spaces.
    A cell denotes a directed acyclic graph of operations (and a motif denotes a small subset of the cell).
    }
    \label{fig:search_space_terms}
\end{figure}

\begin{figure}
    \centering
    \includegraphics[trim=0cm 2cm 2cm  1cm, clip, width=1.0\linewidth]{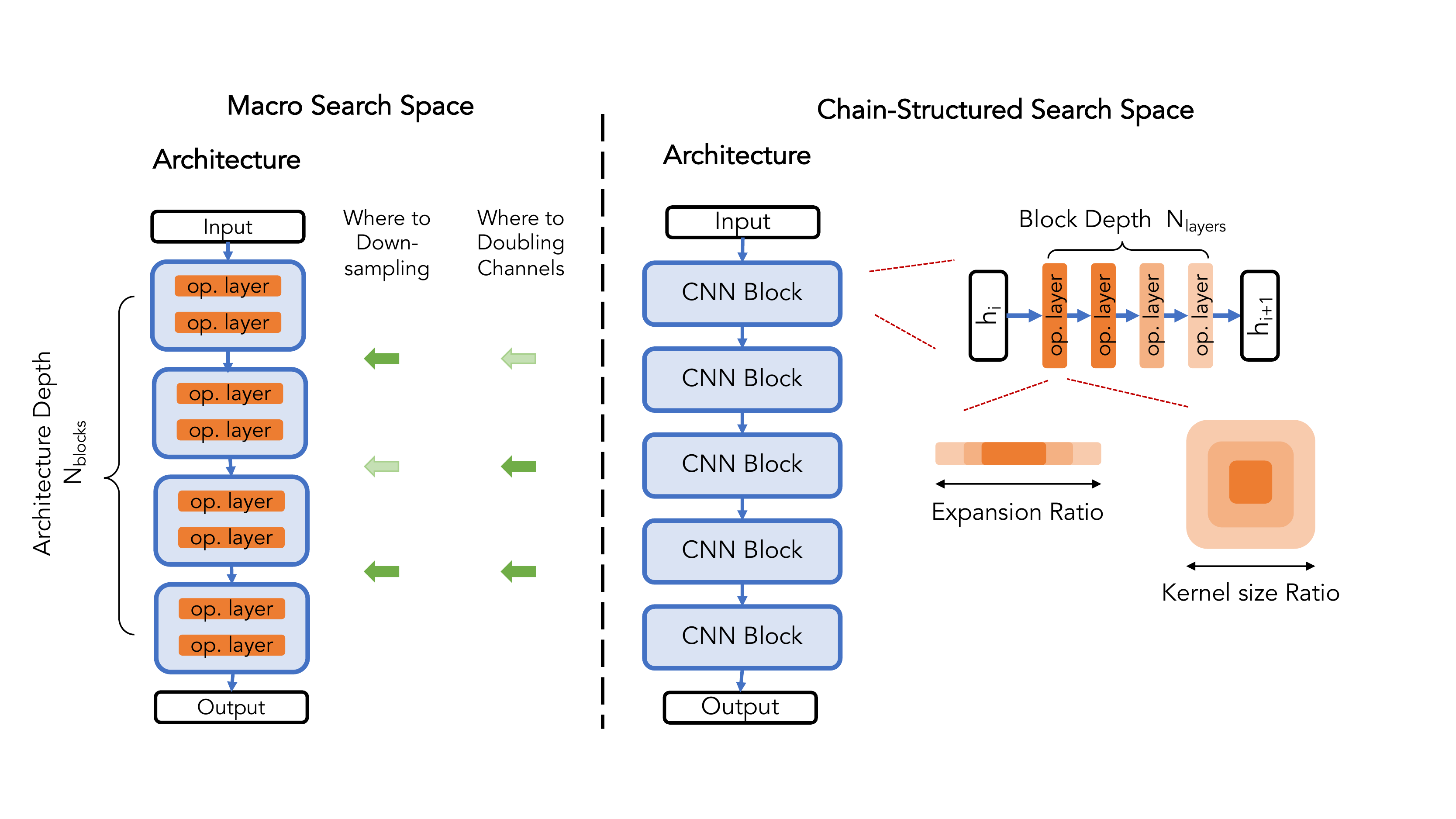}
    \caption{Illustration of macro search space based on \citet{transfernas}(left) and chain-structured search space based on \citet{ofa}(right).}
    \label{fig:chain_macro_ss}
\end{figure}

\begin{figure}
    \centering
    \includegraphics[width=1.0\textwidth]{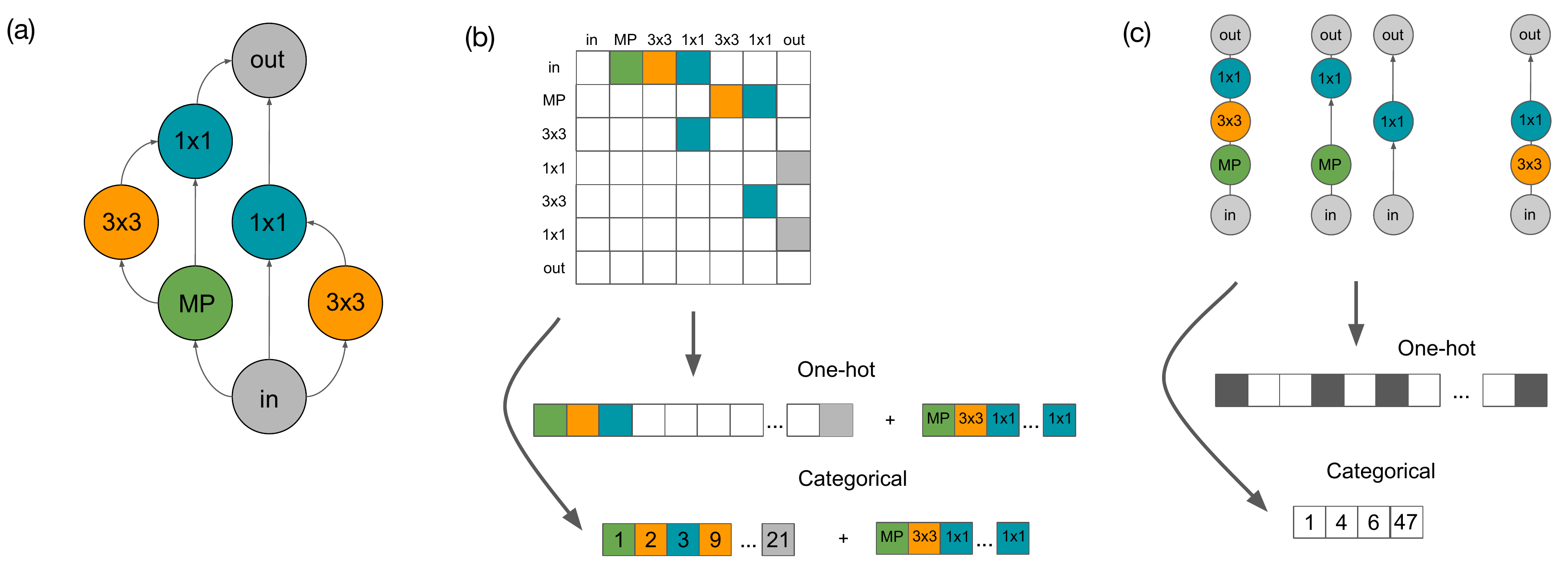} 
    \caption{A neural architecture \emph{(a)} can be encoded using an adjacency matrix \emph{(b)} or path-based representation \emph{(c)}, with a one-hot or categorical encoding.}
    \label{fig:encodings}
\end{figure}

For a visualization of the search space terminologies, see Figure \ref{fig:search_space_terms}.
In Figure \ref{fig:chain_macro_ss}, we show chain-structured and macro search spaces.
Architecture encodings are illustrated in Figure \ref{fig:encodings}.
Finally, for an overview of NAS benchmarks, see Table \ref{tab:benchmarks_comprehensive}.

\begin{table}[t]
\centering
\resizebox{.95\linewidth}{!}{%
\centering
\begin{tabular}{@{}lcccccccc@{}}
\toprule
\multicolumn{1}{l}{} & \multicolumn{1}{c}{} & \multicolumn{1}{c}{} & \multicolumn{2}{c}{\textbf{Queryable}} 
& \multicolumn{1}{c}{} & \multicolumn{1}{c}{} & \multicolumn{1}{c}{} & \multicolumn{1}{c}{} \\
\cmidrule{4-5} \textbf{Benchmark} & \textbf{Size} & \textbf{Type} & \textbf{Tab.} & \textbf{Surr.} 
& \textbf{LCs} & \textbf{One-Shot} & \textbf{Task} & \textbf{\#Tasks} \\
\midrule 
NAS-Bench-101 & 423k & cell & \cmark & & & & Image\ class.\ & 1 \\
\midrule 
NATS-Bench-TSS \\ (NAS-Bench-201) & 6k & cell & \cmark & & \cmark & \cmark & Image\ class.\ & 3 \\
\midrule 
NATS-Bench-SSS & 32k & macro & \cmark & & \cmark &  \cmark & Image\ class.\ & 3 \\
\midrule
NAS-Bench-NLP & $>10^{53}$ & cell & &  & \cmark &  & NLP & 1 \\
\midrule 
NAS-Bench-1Shot1 & 364k & cell & \cmark & & & \cmark & Image\ class.\ & 1 \\
\midrule 
Surr-NAS-Bench-DARTS \\ (NAS-Bench-301) & $10^{18}$ & cell & & \cmark & & \cmark & Image\ class.\ & 1 \\
\midrule
Surr-NAS-Bench-FBNet & $10^{21}$ & chain & & \cmark & & & Image\ class.\ & 1 \\
\midrule
NAS-Bench-ASR & 8k & cell & \cmark & & & \cmark & ASR & 1 \\
\midrule
TransNAS-Bench-101-Micro & 4k & cell & \cmark & & \cmark & \cmark & Var.\ CV & 7 \\
\midrule
TransNAS-Bench-101-Macro & 3k & macro & \cmark & & \cmark & \cmark & Var.\ CV & 7 \\
\midrule
NAS-Bench-111 & 423k & cell & & \cmark & \cmark &  & Image\ class.\ & 1 \\
\midrule
NAS-Bench-311 & $10^{18}$ & cell & & \cmark  & \cmark & \cmark & Image\ class.\ & 1 \\
\midrule
NAS-Bench-NLP11 & $>10^{53}$ & cell & & \cmark & \cmark & & NLP & 1 \\
\midrule
NAS-Bench-MR & $10^{23}$ & cell & & \cmark & & \cmark & Var.\ CV & 9 \\
\midrule
NAS-Bench-Macro & 6k & macro & \cmark & & & \cmark & Image\ class.\ & 1 \\
\midrule
HW-NAS-Bench-201 & 6k & cell & \cmark & &  &  & Image\ class.\ & 3 \\
\midrule
HW-NAS-Bench-FBNet & $10^{21}$ & chain & \cmark & & & & Image\ class.\ & 1 \\
\midrule
\midrule
NAS-Bench-360 & Var.\ & suite & \cmark & & \cmark & \cmark & Var.\ & 3 \\
\midrule
NAS-Bench-Suite & Var.\ & suite & \cmark & \cmark & \cmark & \cmark & Var.\ & 25 \\
NAS-Bench-Suite-Zero & Var.\ & suite & \cmark & \cmark & \cmark & \cmark & Var.\ & 28 \\
\bottomrule
\end{tabular}
}
\caption{An overview of NAS benchmarks.
\label{tab:benchmarks_comprehensive}
}
\end{table}




\clearpage
\bibliography{main}

\end{document}